\documentclass[pdflatex, sn-nature]{sn-jnl}

\usepackage{graphicx}
\graphicspath{{./}{../}}
\usepackage{multirow}
\usepackage{amsmath,amssymb,amsfonts,mathtools}
\usepackage{amsthm}
\usepackage{mathrsfs}
\usepackage[title]{appendix}
\usepackage{xcolor}
\usepackage{textcomp}
\usepackage{manyfoot}
\usepackage{booktabs}
\usepackage{algorithm}
\usepackage{algorithmicx}
\usepackage{algpseudocode}
\usepackage{listings}
\usepackage{cases}
\usepackage{microtype}
\usepackage{hyperref}
\usepackage{lineno}

\newcommand{\Var}{\mathrm{Var}}
\newcommand{\Cov}{\mathrm{Cov}}

\newcommand{\dd}{\textup{d}}
\newcommand{\bv}{\mathbf{d}_v}
\newcommand{\bh}{\mathbf{d}_h}
\newcommand{\sigm}{\mathrm{sigmoid}}

\newcommand{\panellabel}[1]{\textbf{#1}}
\newcommand{\panelrow}[1]{\makebox[\linewidth][l]{\panellabel{#1}}\\[0.5pt]}
\newcommand{\etal}{\textit{et~al.}}

\theoremstyle{thmstyleone}

\begin{document}

%\linenumbers

\title[CVSI for Diffusion Models]{Control Variate Score Matching for Diffusion Models}

\author[1,2,3]{\fnm{Khaled} \sur{Kahouli}}
\author[1]{\fnm{Romuald} \sur{Elie}}
\author*[1,2,3,4,5]{\fnm{Klaus-Robert} \sur{Müller}}\email{klaus-robert.mueller@tu-berlin.de}
\author[1]{\fnm{Quentin} \sur{Berthet}}
\author*[1]{\fnm{Oliver T.} \sur{Unke}}\email{oliverunke@google.com}
\author*[1]{\fnm{Arnaud} \sur{Doucet}}\email{arnauddoucet@google.com}

\affil[1]{\orgname{Google DeepMind and Google Research}}
\affil[2]{\orgname{BIFOLD – Berlin Institute for the Foundations of Learning and Data}}
\affil[3]{\orgdiv{Machine Learning Group}, \orgname{Technische Universit\"at Berlin}}
\affil[4]{\orgdiv{Department of Artificial Intelligence}, \orgname{Korea University}}
\affil[5]{\orgname{Max-Planck Institute for Informatics}}

\abstract{
Sampling from unnormalized probability densities is a pervasive challenge across the computational and physical sciences. Diffusion models provide a powerful generative framework for this task, but their success relies on accurately estimating the score of the perturbed target distribution. Current approaches face a dichotomy between two standard estimation methods: the Denoising Score Identity (DSI) requires data samples and exhibits high variance at low noise levels, whereas the Target Score Identity (TSI) relies on the energy function and suffers from diverging variance at high noise levels. In this work, we reconcile both approaches by introducing the Control Variate Score Identity (CVSI), an unbiased estimator with an analytically optimal, state- and time-dependent control coefficient that theoretically minimizes variance over the entire diffusion process. CVSI serves as a robust plug-in estimator that significantly enhances performance and efficiency in data-free sampler learning and training-free diffusion sampling. These gains scale to complex, high-dimensional energy-based models.
}


\maketitle

%\section{Introduction}
\label{intro}

\begin{figure}[!tp]
    \centering
    \begin{minipage}{1.0\linewidth}
        \centering
        \begin{minipage}[t]{0.49\linewidth}
            \panelrow{a}
            \centering
            \includegraphics[width=\linewidth]{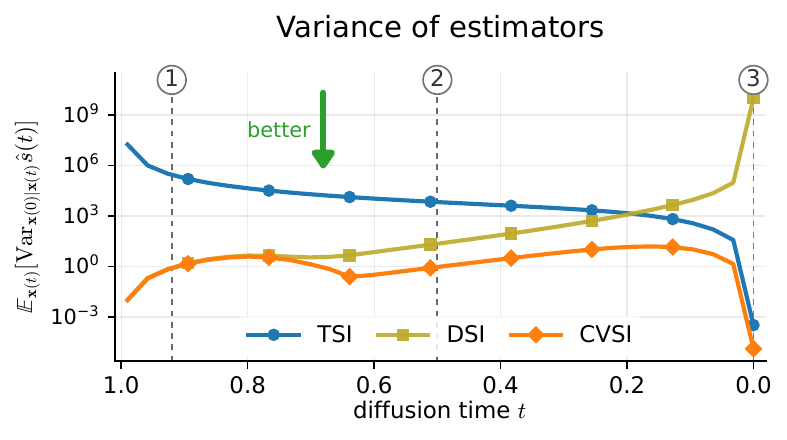}
        \end{minipage}
        \hfill
        \begin{minipage}[t]{0.49\linewidth}
            \panelrow{b}
            \centering
            \includegraphics[width=\linewidth]{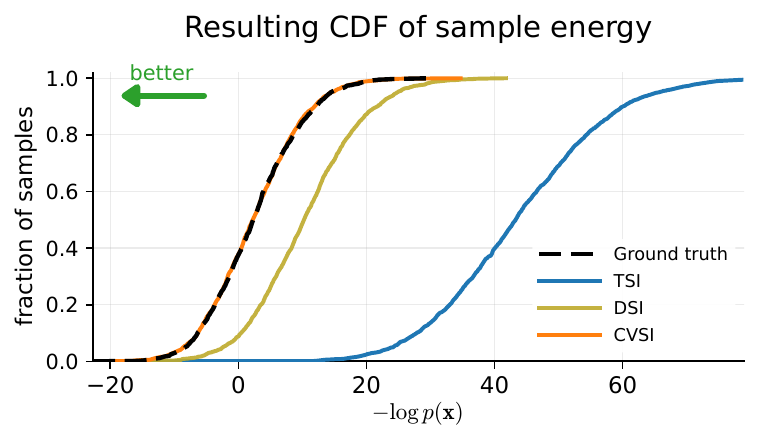}
        \end{minipage}
    \end{minipage}

    \begin{minipage}{1.0\linewidth}
        \centering
        \begin{minipage}[t]{0.590\linewidth}
            \panelrow{c}
            \centering
            \includegraphics[width=\linewidth]{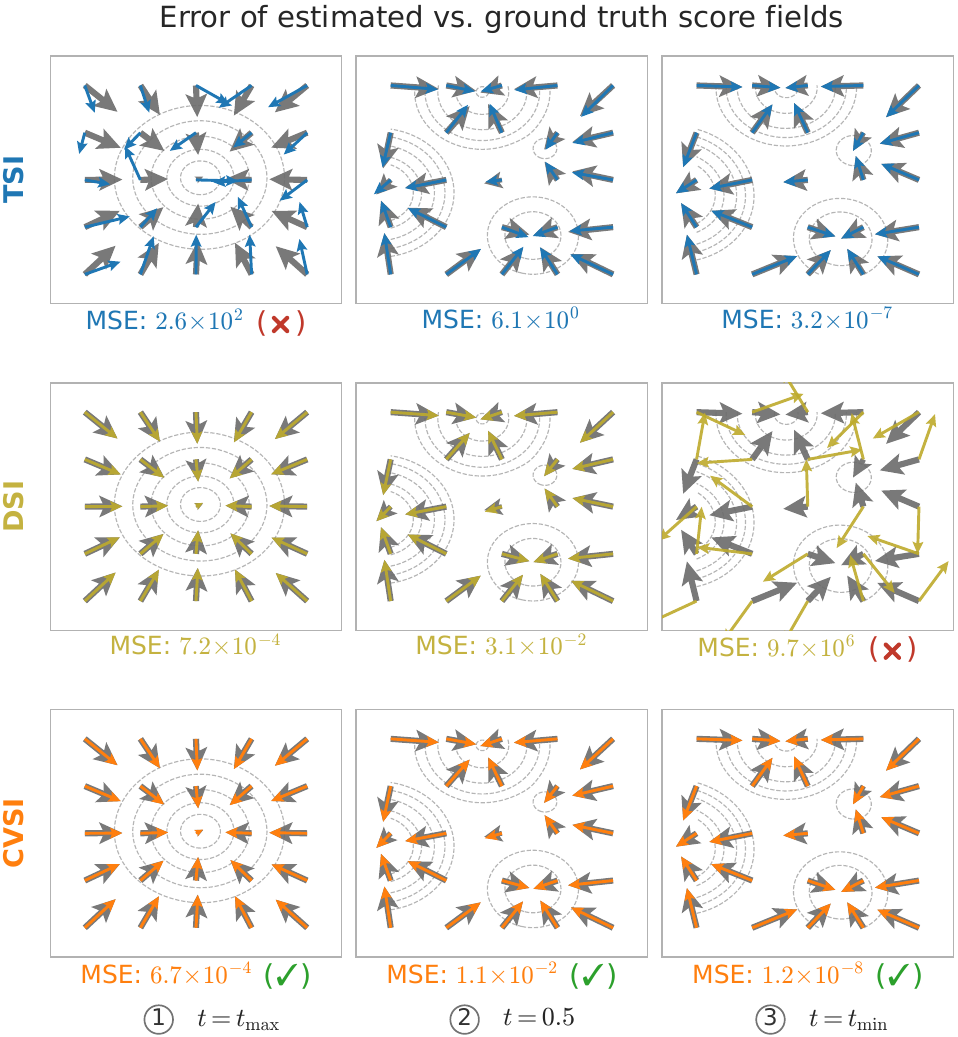}
        \end{minipage}
        %\hfill
        \begin{minipage}[t]{0.185\linewidth}
            \panelrow{d}
            \centering
            \includegraphics[width=\linewidth]{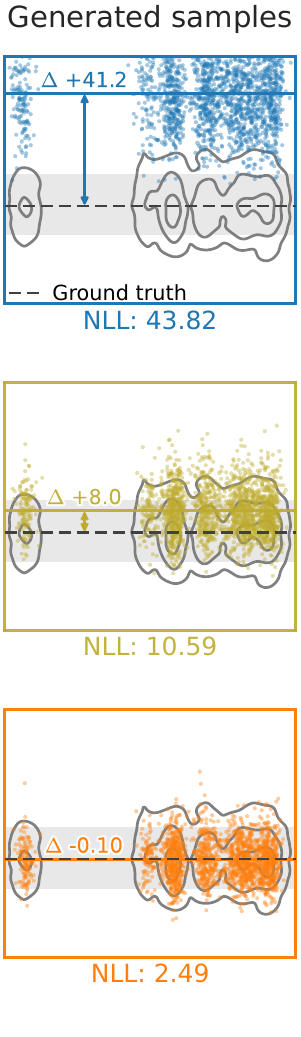}
        \end{minipage}
    \end{minipage}
    \caption{
        \textbf{Variance reduction via control variates.}
        All panels evaluate a $D=100$, 20-component GMM using a VP-ISSNR diffusion schedule ($\eta=1.0$, $\kappa=0$) \cite{kahouli2025tv_snr}.
        \textbf{a,} Estimator variance (log scale) as a function of diffusion time $t$. DSI and TSI diverge at $t \to 0$ and $t \to 1$, respectively. Our CVSI (orange) consistently achieves the theoretical minimum variance, staying orders of magnitude lower than the baselines in the critical regimes. CVSI's interpolation weight $\tilde{c}$ is shown in Supplementary Fig.~\ref{fig:tilde_ct}.
        \textbf{b,} Empirical cumulative distribution function (CDF) of the sampled energies, $-\log p(\mathbf{x})$. While TSI generates samples almost entirely off the target manifold and DSI exhibits a high-energy tail, CVSI accurately recovers the ground truth (GT).
        \textbf{c,} The estimated score fields at the three representative diffusion times marked in \textbf{a}, projected onto the leading two principal components of the target. The analytic GT field is marked by grey arrows, and the dashed contours represent its marginal $q_t$ (the resulting MSE is shown below each panel). 
        %The resulting MSE is shown below each plot, where green check marks highlight the best estimator at each diffusion time and red crosses represent severely failing cases.
        Because all three estimators are theoretically unbiased, their empirical MSE is dominated entirely by their variance. Consequently, TSI and DSI suffer from large magnitude and directional errors at opposing ends of the diffusion process, while CVSI accurately recovers the analytic GT at all noise levels.
        \textbf{d,} The energy, $-\log p(\mathbf{x})$, of generated samples against their projection onto the leading principal component. The grey contours mark the target distribution, and the shaded gray band marks the optimal energy range. A dashed line marks the GT average NLL and a solid line marks the estimator's average NLL, with their signed difference $\Delta$. CVSI is the only estimator whose samples and resulting NLL align accurately with the GT ($\Delta = -0.10$).
    }
    \label{fig:control_variate_variance}
\end{figure}
Sampling from unnormalized probability densities $p(\mathbf{x}) \propto e^{-E(\mathbf{x})}$, where the underlying energy function $E(\mathbf{x})$ of a system is known but the normalization constant is intractable, remains a fundamental and longstanding challenge in the physical and computational sciences. In domains ranging from lattice field theory to molecular dynamics and proteins \cite{noe2019boltzmann, albergo2019flow, nicoli2020asymptotically, Jumper2021, Qiao2024, bioemu}, important observables can often be computed by sampling representative states from complex, multimodal energy landscapes. Traditional computational approaches, such as Markov chain Monte Carlo sampling (MCMC) \cite{DoucetFG01, Andrieu2003}, often struggle due to the high dimensionality inherent in these systems, a bottleneck that has driven the search for more efficient and scalable sampling paradigms \cite{Roberts_mcmc, neal2011mcmc, CotterMCMC}.

Recently, diffusion models \cite{diff_mod_sohl, DDPM_Ho, song2021score} have emerged as a powerful generative framework to overcome this challenge \cite{zhang2021path, midgley2022flow, vargas2023denoising, berner2024an, sendera2024improved, phillips2024particle, grenioux2024stochastic, huang2023reverse, akhound2024iterated, akhoundsadegh2025progressive,  choi2025noneqsampler, liu2025adjointschrodingerbridgesampler, adjointsampling}. By simulating a forward stochastic process $\mathbf{x}(t)$ that gradually corrupts data over time $t \in [0,1]$ into unstructured noise, and subsequently learning to reverse this trajectory, diffusion models can map simple prior distributions $q_{t=1}(\mathbf{x})$ onto complex target densities $q_{t=0}(\mathbf{x})=p(\mathbf{x})$. The quantity driving this reverse generative process is the score function, the spatial gradient of the log-probability density of the perturbed data, i.e., $\nabla_{\mathbf{x}(t)} \log q_t(\mathbf{x}(t))$. However, this score function is usually analytically intractable, and accurately and efficiently estimating it represents the critical computational bottleneck in diffusion-based sampling.

The standard approach to estimating the score function is the Denoising Score Identity (DSI) \cite{vincent}, which requires access to target training samples. However, DSI exhibits a severe limitation: it suffers from increasingly high variance near the data manifold at low noise levels ($t \to 0$). In scientific domains where the unnormalized target energy function is accessible, this reliance on pre-computed data can be bypassed entirely by instead estimating the score function via the Target Score Identity (TSI) \cite{de2024target}. Yet, while TSI successfully eliminates the low-noise instability of DSI, it suffers from an opposing failure mode: its variance grows rapidly at high noise levels ($t \to 1$) where the correlation between the perturbed state and the clean target vanishes (Fig.~\ref{fig:control_variate_variance}a). Together, these two identities leave score-based sampling with a stark bias-variance trade-off (cf.~\cite{geman1992neural}).

Current data-free sampling algorithms approximate the score by minimizing the reverse Kullback–Leibler divergence over path measures \cite{zhang2021path, berner2024an, vargas2023denoising}, employing computationally expensive MCMC approaches guided by DSI \cite{huang2023reverse,grenioux2024stochastic}, or using importance sampling approximations guided by TSI \cite{akhound2024iterated}. To mitigate the high variance inherent to either identity, alternative formulations propose hybrid regression targets that explicitly combine both \cite{de2024target}, most notably the concurrent work by Ko \etal~\cite{ko2025latent}. Adjoint sampling algorithms \cite{adjointsampling} also implicitly exploit a linear combination of such regression targets. However, these strategies largely rely on heuristic weighting schedules or lack a rigorous theoretical justification for variance minimization. Consequently, they result in suboptimal solutions, which limits their effectiveness. Moreover, their ability to reduce variance in MCMC estimation regimes remains unclear.

In this work, we rigorously alleviate this trade-off by unifying both score identities within the principled statistical framework of control variates \cite{lemieux2014control}. To this end, we introduce the Control Variate Score Identity (CVSI), an unbiased generalized estimator that features an analytically optimal, state- and time-dependent control coefficient. By dynamically interpolating between DSI and TSI without introducing any bias, CVSI is theoretically guaranteed to minimize the variance across the entire noise spectrum, and consequently reduce the resulting mean squared error (MSE) (Fig.~\ref{fig:control_variate_variance}a,c). We demonstrate empirically that it serves as a robust plug-in replacement for standard estimators and eliminates the need for heuristic tuning. Because CVSI strictly minimizes estimator variance, it significantly reduces the required number of expensive energy evaluations, resulting in substantial performance improvements under limited computational budgets. Consequently, our CVSI significantly reduces sample complexity and improves sampling performance across two primary regimes: training-free diffusion sampling using MCMC methods and data-free sampler learning, the latter of which entirely bypasses the need for large pre-computed datasets. Finally, we demonstrate that these efficiency gains scale to complex applications by validating CVSI on a high-dimensional image energy-based model, achieved by deriving an exact augmented diffusion posterior for a Gaussian--Bernoulli restricted Boltzmann machine.

\section{Results}\label{results}

\subsection{The control variate score identity (CVSI)}
\paragraph{Unified control variate estimator.}
Score-based sampling is limited by a fundamental dichotomy between the two standard estimation identities of the marginal distribution score, $\nabla_{\mathbf{x}(t)} \log q_t(\mathbf{x}(t))$. Given a standard forward diffusion process with perturbation kernel $q(\mathbf{x}(t)|\mathbf{x}(0)) = \mathcal{N}(\mathbf{x}(t); a(t)\mathbf{x}(0), b(t)^2\mathbf{I})$, the Denoising Score Identity (DSI), which leverages the conditional score $\nabla_{\mathbf{x}(t)} \log q(\mathbf{x}(t)|\mathbf{x}(0))$, exhibits high variance near the data manifold ($t \to 0$), whereas the Target Score Identity (TSI), relying on the clean data score $\nabla_{\mathbf{x}(0)} \log p(\mathbf{x}(0))$, suffers from variance divergence at high noise levels ($t \to 1$). In this work, we rigorously resolve this trade-off by deriving a unified score function estimator that strictly minimizes variance across the entire diffusion process.

To this end, we reformulate the score estimation within the framework of \textit{control variates}, where we use the score of the posterior distribution $\nabla_{\mathbf{x}(0)} \log q(\mathbf{x}(0) | \mathbf{x}(t))$ as a control variate for the standard TSI estimator. They are linearly related via Bayes' rule, ensuring high correlation, while standard regularity conditions guarantee that $\mathbb{E}_{q(\mathbf{x}(0) | \mathbf{x}(t))} [\nabla_{\mathbf{x}(0)} \log q(\mathbf{x}(0) | \mathbf{x}(t))] = 0$. This results in the \textbf{Control Variate Score Identity (CVSI)}, a family of unbiased estimators $\hat{\mathbf{s}}_\mathrm{CV}$ for $\nabla_{\mathbf{x}(t)} \log q_t(\mathbf{x}(t))$ governed by a state- and time-dependent control coefficient $c(\mathbf{x}(t), t) \in \mathbb{R}$:
\begin{equation}
\begin{split}
    \hat{\mathbf{s}}_\mathrm{CV}(\mathbf{x}(t)) &= \mathbb{E}_{q(\mathbf{x}(0) | \mathbf{x}(t))} \bigg[ \frac{1}{a(t)} \nabla_{\mathbf{x}(0)} \log p(\mathbf{x}(0))
     - c(\mathbf{x}(t), t) \nabla_{\mathbf{x}(0)} \log q(\mathbf{x}(0) | \mathbf{x}(t)) \bigg].
\end{split}
\label{eq:cvsi_result}
\end{equation}

\paragraph{Optimal control coefficient.}
While the CVSI is unbiased for any choice of the scalar function $c(\mathbf{x}(t), t)$, we derive the optimal $c^*(\mathbf{x}(t), t)$ that strictly minimizes the estimator's variance. It depends only on the target score $\mathbf{s}_p = \nabla_{\mathbf{x}(0)} \log p(\mathbf{x}(0))$ and the analytical perturbation kernel score $\mathbf{s}_{t|0} = \nabla_{\mathbf{x}(t)} \log q(\mathbf{x}(t)|\mathbf{x}(0))$ and can be tractably computed as:
\begin{equation}
\begin{split}
    c^*(\mathbf{x}(t), t) = \frac{\Var(\mathbf{s}_p) - a(t) \Cov(\mathbf{s}_p, \mathbf{s}_{t|0})}
     {a(t) \Var(\mathbf{s}_p) + a(t)^3 \Var(\mathbf{s}_{t|0}) - 2 a(t)^2 \Cov(\mathbf{s}_p, \mathbf{s}_{t|0})},
\end{split}
\label{eq:opt_c_t_result}
\end{equation}
where all statistics are taken w.r.t. the posterior $q(\mathbf{x}(0) | \mathbf{x}(t))$, for a given state $\mathbf{x}(t)$. This optimal coefficient dynamically adjusts the contribution of the control variate based on the instantaneous correlation between the target geometry and the diffusion posterior, effectively reducing the variance throughout the full noise spectrum. This is in contrast to standard methods like DSI and TSI, which suffer from variance divergence at opposing noise levels (Fig.~\ref{fig:control_variate_variance}a,c).

\paragraph{Unifying previous score identities.}
Crucially, our formulation unifies DSI and TSI as endpoints of a single continuum. By defining the mixing weight $\tilde{c}(\mathbf{x}(t), t) = c(\mathbf{x}(t), t) \, a(t)$, Eq.~\eqref{eq:cvsi_result} acts as a convex interpolation of both identities:
\begin{align}
    \hat{\mathbf{s}}_\mathrm{CV}(\mathbf{x}(t)) & = \frac{1 - \tilde{c}(\mathbf{x}(t), t)}{a(t)} \, \underbrace{\mathbb{E}_{q_{0|t}} \left[ \nabla \log p(\mathbf{x}(0)) \right]}_{\text{TSI Component}} \nonumber \\
     &\quad + \tilde{c}(\mathbf{x}(t), t) \, \underbrace{\mathbb{E}_{q_{0|t}} \left[ \nabla \log q(\mathbf{x}(t) | \mathbf{x}(0)) \right]}_{\text{DSI Component}}.
     \label{eq:interpolated_result}
\end{align}
This reveals that TSI ($\tilde{c}=0$), DSI ($\tilde{c}=1$), and heuristic mixtures like TSM~\cite{de2024target} (Supplementary Table~\ref{tab:estimators_comparison} in Section~\ref{app:unif_prev}) are merely special, suboptimal cases of our framework. In contrast, our theoretically optimal weighting function $\tilde{c}^*(\mathbf{x}(t), t) = c^*(\mathbf{x}(t), t)a(t)$ depends on the data state and the noise schedule and automatically minimizes variance by transitioning smoothly from $0$ to $1$ as diffusion time progresses (Supplementary Fig.~\ref{fig:tilde_ct}), without requiring manual tuning or ground truth knowledge.

\subsection{Empirical validation}

\begin{figure*}[t]
    \centering
    \begin{minipage}{1.0\linewidth}
        \centering
        \begin{minipage}[t]{0.58\linewidth}
            \panelrow{a}
            \centering
            \begin{minipage}{0.48\linewidth}
                \centering
                \includegraphics[width=\linewidth]{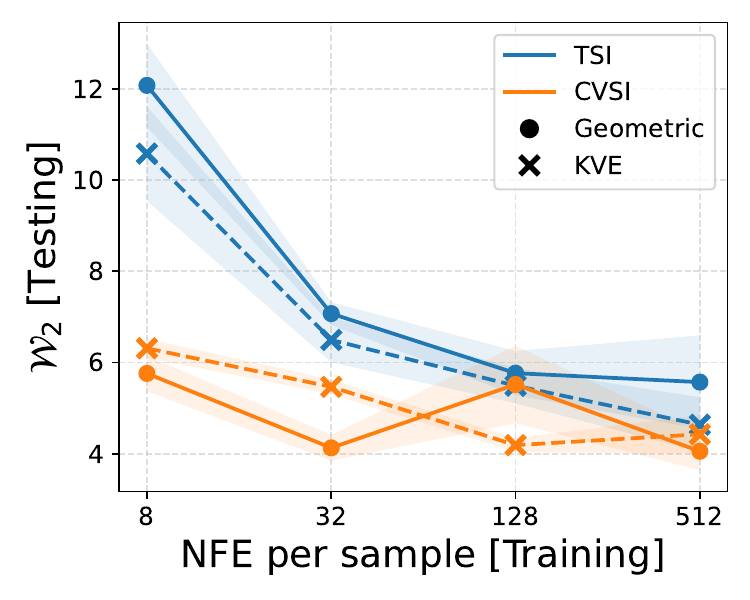}
            \end{minipage}
            \hfill
            \begin{minipage}{0.48\linewidth}
                \centering
                \includegraphics[width=\linewidth]{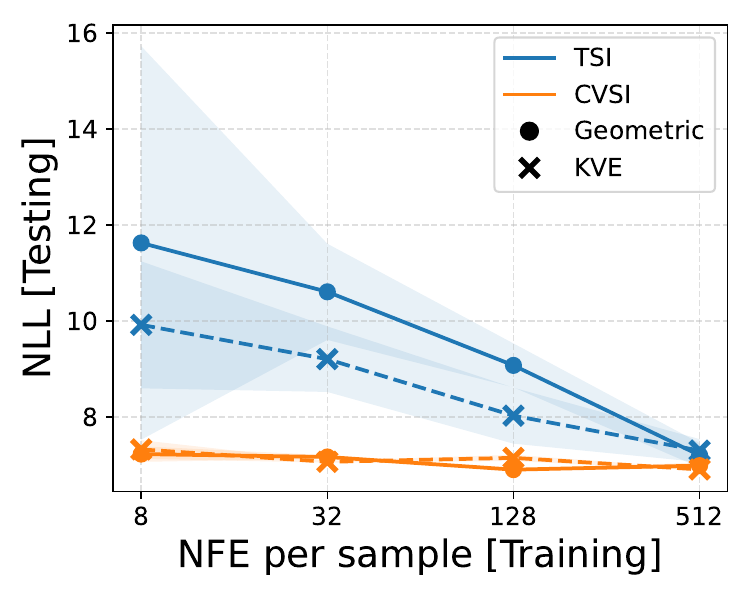}
            \end{minipage}
        \end{minipage}
        \hfill
        \begin{minipage}[t]{0.4\linewidth}
            \panelrow{b}
            \centering
            \begin{minipage}{0.48\linewidth}
                \centering
                \includegraphics[width=\linewidth]{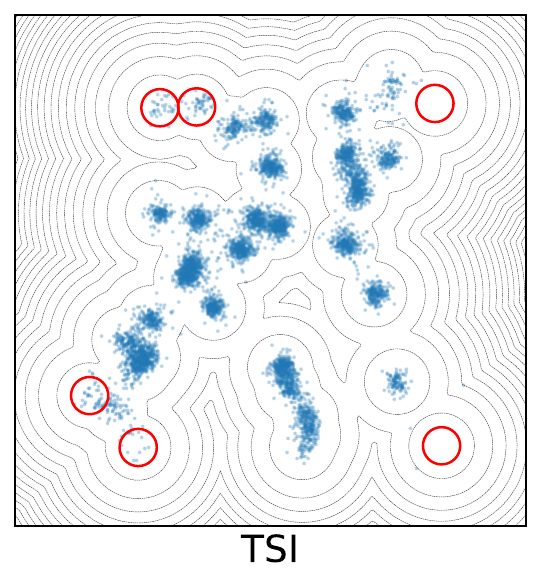}
                %{\footnotesize iDEM+TSI (8 NFE)}
            \end{minipage}
            \hfill
            \begin{minipage}{0.48\linewidth}
                \centering
                \includegraphics[width=\linewidth]{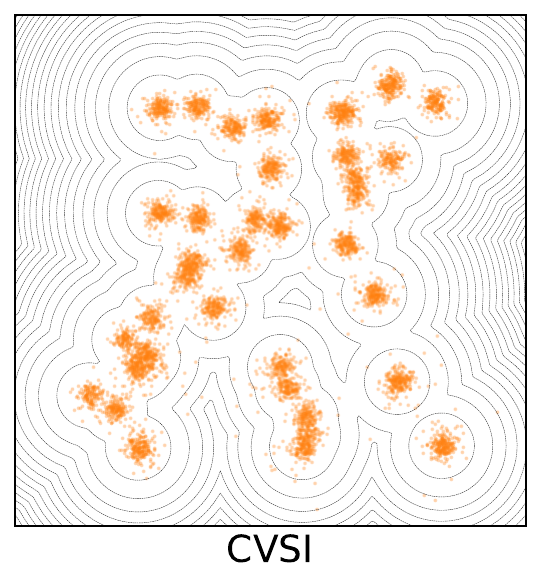}
                %{\footnotesize iDEM+CVSI (8 NFE)}
            \end{minipage}
        \end{minipage}
    \end{minipage}

    \vspace{0.2cm}

    \begin{minipage}{1.0\linewidth}
        \centering
        \begin{minipage}[t]{0.32\linewidth}
            \panelrow{c}
            \centering
            \includegraphics[width=1.\linewidth]{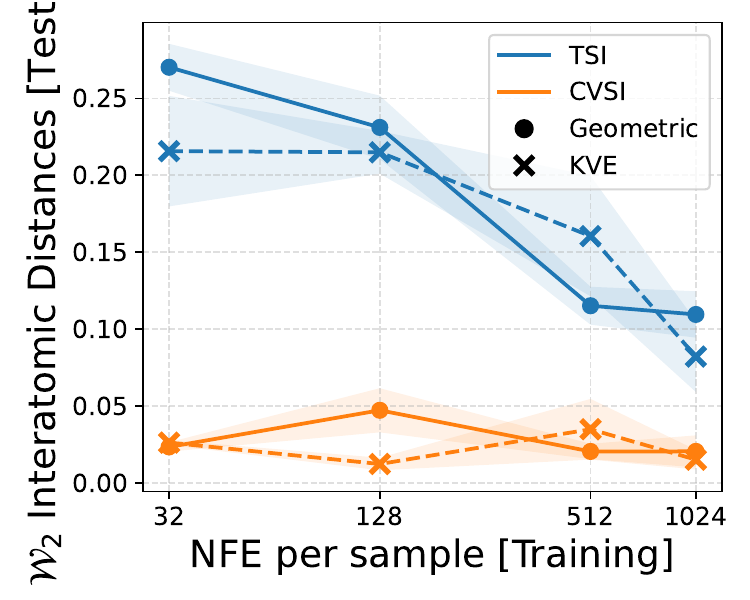}
        \end{minipage}
        \hfill
        \begin{minipage}[t]{0.64\linewidth}
            \panelrow{d}
            \centering
            \begin{minipage}{0.48\linewidth}
                \centering
                \includegraphics[width=\linewidth]{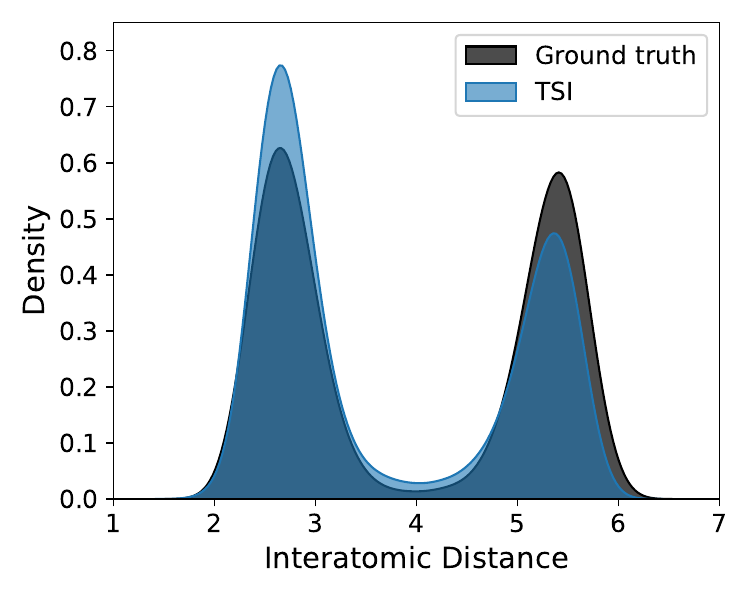}
                %{\vspace{0.2cm} \footnotesize iDEM+TSI (128 NFE)}
            \end{minipage}
            \hfill
            \begin{minipage}{0.48\linewidth}
                \centering
                \includegraphics[width=\linewidth]{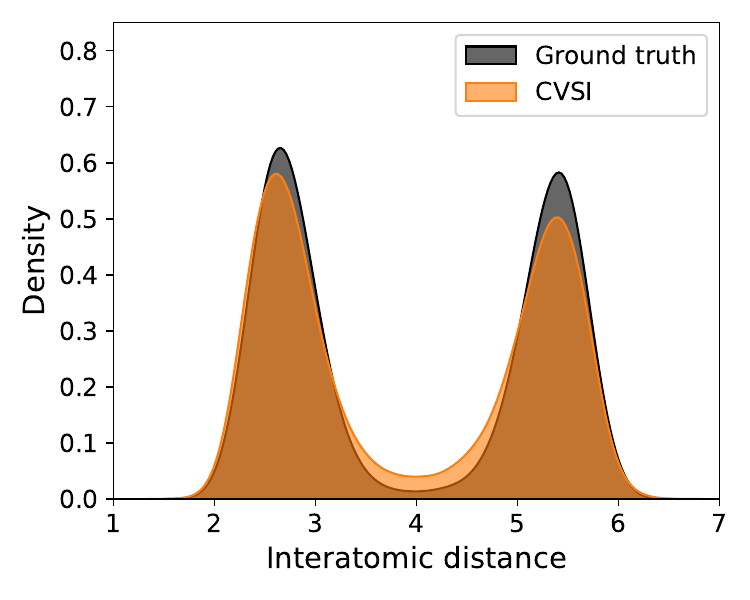}
                %{\vspace{0.2cm} \footnotesize iDEM+CVSI (128 NFE)}
            \end{minipage}
        \end{minipage}
    \end{minipage}
    \caption{
    \textbf{Data-free sampler learning with CVSI.}
    Comparison of standard iDEM using TSI versus our CVSI across two energy landscapes. Markers indicate the training noise schedule (circles: Geometric, crosses: KVE).
    \textbf{{a,}} 40-Component 2D GMM: CVSI achieves lower Wasserstein-2 distance ($\mathcal{W}_2$) and Negative Log-Likelihood (NLL) with significantly fewer energy function evaluations (NFEs) per training sample. \textbf{b,} The resulting GMM distribution plots, where the ground truth is shown as contour lines, reveal that at a budget of only 8 NFEs, TSI suffers from severe mode dropping (highlighted with red circles), whereas CVSI successfully covers all target modes.
    \textbf{c,} 2D 4-Particle Double-Well Potential: Consistent with the GMM case, CVSI results in a lower $\mathcal{W}_2$ error for interatomic distances across all tested NFE budgets. \textbf{d,} The histograms of interatomic distances confirm that using a budget of 128 NFEs, CVSI recovers the ground truth distribution more accurately, whereas TSI incorrectly oversamples short distances.
    }
    \label{fig:idem_combined_results}
\end{figure*}

We empirically validate our CVSI across two distinct regimes: training-free diffusion sampling and data-free sampler learning, where estimator variance dictates sample efficiency and scalability to high dimensions.

\paragraph{Training-free diffusion sampling.}
We begin our evaluation by assessing the performance and efficiency of CVSI on Gaussian Mixture Models (GMMs). We select this setting because the posterior $q(\mathbf{x}(0)|\mathbf{x}(t))$ admits a closed-form analytical solution. Although a closed-form analytical solution is not required (we can use MCMC to draw samples from the posterior~\cite{huang2023reverse}), this setting allows us to rigorously isolate the variance of the estimators from posterior approximation errors. At the same time it enables us to control the system's dimensionality and complexity which helps to carefully study challenging high-dimensional distributions while having access to the ground truth.

We find that standard estimators (DSI, TSI) and heuristic mixing strategies (e.g., TSM) degrade rapidly as the system dimensionality and the target distribution complexity increase (Supplementary Fig.~\ref{fig:gmm_nll_vs_dim_comp}). In contrast, CVSI maintains Negative Log-Likelihood (NLL) values nearly indistinguishable from the ground truth, as demonstrated on a 20-component GMM in $D=100$ in Fig.~\ref{fig:control_variate_variance}b,d. Crucially, this stability translates into a significant reduction in computational cost. CVSI achieves optimal performance with as few as 5 MC posterior samples per step, whereas baselines require up to 6$\times$ more samples to achieve comparable results (see Supplementary Fig.~\ref{fig:gmm_nll_vs_dim_comp} for a detailed overview). Because each MC sample requires the evaluation of the target energy function, this increase in sample efficiency translates to a speedup in sampling time. This is crucial for computationally expensive systems, and makes previously prohibitive training-free sampling more feasible.

\paragraph{Data-free sampler learning.}

Beyond training-free sampling, the low-variance properties of CVSI translate to improved training dynamics in data-free sampler learning. In this setting, the model must learn purely from the unnormalized energy function $E(\mathbf{x})$ without access to data samples. To this end, we integrate CVSI into the iDEM algorithm~\cite{akhound2024iterated}, substituting their standard TSI estimator, and demonstrate it on a 40-component 2D GMM and a 2D 4-particle Double-Well (DW-4) potential.

As illustrated in Fig.~\ref{fig:idem_combined_results}, CVSI produces significantly higher sample quality than the baseline across varying noise schedules (Geometric~\cite{song2021score} and KVE~\cite{diff_edm}), evidenced by lower Negative Log-Likelihood (NLL) and Wasserstein-2 ($\mathcal{W}_2$) distances. This advantage is decisive in computationally constrained regimes. Given a strictly limited budget, as low as 8 NFEs per training sample, the baseline model fails to accurately capture the target distribution, resulting in severe mode dropping for the GMM and a distorted distribution in the DW-4 system where short interatomic distances are oversampled. In contrast, using the same minimal computational budget, CVSI successfully resolves all modes and more faithfully recovers the target distribution by sufficiently lowering the variance. This data efficiency unlocks the ability to train samplers directly from the energy function without requiring a pre-computed dataset. 
This is particularly crucial when obtaining diverse, unbiased data a priori is computationally expensive or inherently elusive, and training on suboptimally generated data could fundamentally bias the model.

Unlike in training-free diffusion sampling, scaling the target distribution to higher dimensions, e.g. a 6-dimensional GMM, with iDEM results in instability during training and degraded performance. In addition to the amplified variance of importance weights in high dimensions, our theoretical analysis reveals that score approximation errors during early training artificially inflate the sampling temperature when discretizing the SDE. This generates overly noisy replay buffer samples, resulting in a deteriorating feedback loop that degrades subsequent training steps. To counteract this stochasticity induced by the error, we decouple the diffusion noise scale from the drift, introducing an independent parameter $\lambda_\mathrm{eff}$ for the diffusion term to reduce noise injection. While  standard TSI still suffers from severe mode collapse under these conditions (see Supplementary Fig.~\ref{fig:gmm6d_lambda}), our low-variance CVSI, combined with a reduced $\lambda_\mathrm{eff} < 1$, stabilizes the training loop and mitigates the temperature inflation, faithfully recovering the distribution modes.

\subsection{Real-world application}
\label{sec:real_world}

The DW-4 and GMM benchmarks above serve to empirically validate our framework under controlled settings, with the GMM benchmark allowing direct control over the dimensionality and complexity of the target, effectively isolating the estimator variance in the case of training-free sampling, where the posterior is analytically known. We now extend our evaluation to high-dimensional and complex energy landscapes that are more representative of real-world data. 

\paragraph{Sampling from high-dimensional image energy models.}
\label{sec:real_world_images}

\begin{figure*}[t]
    \centering
    % Assembled from the three letterless panels written by
    % iDEM/mnist_plots.py (fig_main_panels), so \panellabel sets the letters and
    % they match Figs. 1 and 2 exactly. Widths keep the 1.75 : 1 : 1 ratio of the
    % previous single-file mnist_main.pdf, which is still written as a preview.
    \begin{minipage}[t]{0.50\linewidth}
        \panelrow{a}
        \centering
        \includegraphics[width=\linewidth]{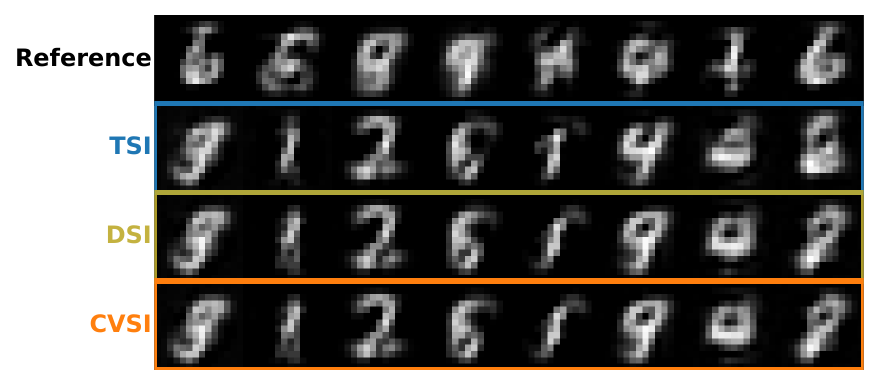}
    \end{minipage}
    \hfill
    \begin{minipage}[t]{0.235\linewidth}
        \panelrow{b}
        \centering
        \includegraphics[width=\linewidth]{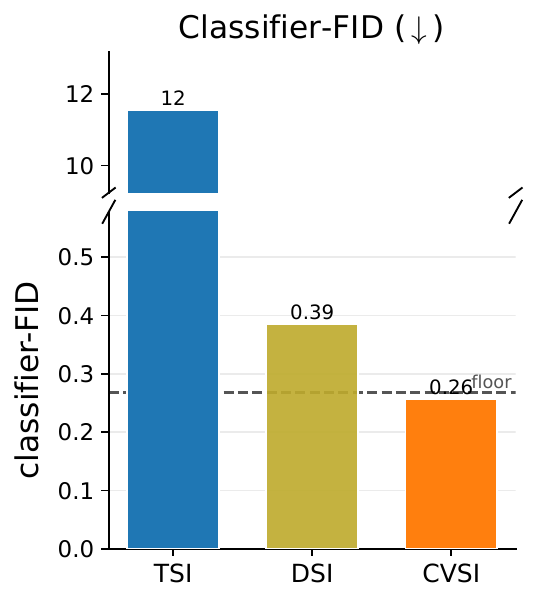}
    \end{minipage}
    \hfill
    \begin{minipage}[t]{0.235\linewidth}
        \panelrow{c}
        \centering
        \includegraphics[width=\linewidth]{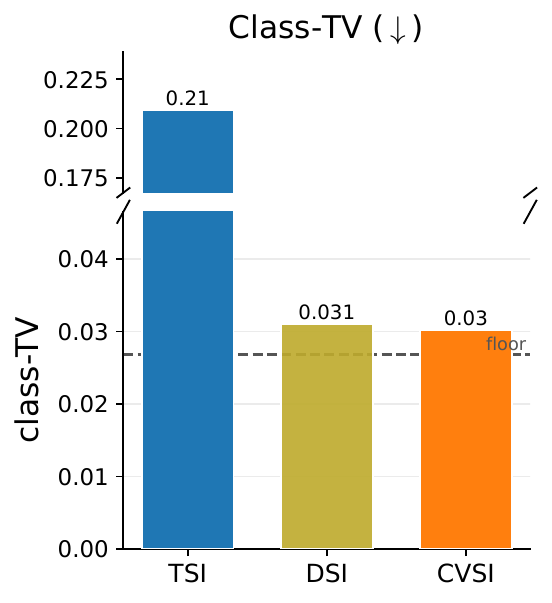}
    \end{minipage}
    \caption{
    \textbf{Sampling from a Gaussian--Bernoulli RBM energy model for images.}
    \textbf{a,} Eight digits generated by simulating the reverse SDE with each estimator, next to reference samples drawn from the learned energy model $p_\theta(\mathbf{x})$ using a long-run block-Gibbs sampler.
    \textbf{b,} Classifier-FID and \textbf{c,} class-TV against the reference samples, both at the smallest Monte Carlo budget $K=2$. The dashed line marks the reference floor, and the $y$-axis is split due to large TSI values. CVSI is closest to the reference floor performance.
    }
    \label{fig:mnist_results}
\end{figure*}

We use a Gaussian--Bernoulli restricted Boltzmann machine (GB--RBM) trained on $14\times14$ MNIST digits  ($D=196$). This provides an unnormalized target density, $p(\mathbf{x})\propto\exp(-F(\mathbf{x}))$, that captures the complex multimodal structure of images, which otherwise do not possess an analytic energy landscape. Crucially, we demonstrate that the bipartite structure of the GB--RBM extends to the diffusion framework  (see Online Methods~\ref{sec:gbrbm_exact_sampling}). This provides closed-form conditionals that allow the diffusion posterior, $q(\mathbf{x}(0)\vert{}\mathbf{x}(t))$, to be sampled via block-Gibbs iterations. This guarantees asymptotically exact and unbiased posterior sampling, ensuring that performance discrepancies between TSI, DSI, and CVSI stem strictly from intrinsic estimator variance, now scaled to a $196$-dimensional space. 

We generate digits by simulating the reverse SDE (Eq.~\eqref{eq:methods_rev_sde}), estimating the score at each integration step using short-run block-Gibbs posterior draws. To rigorously quantify generative fidelity and mode balance, we evaluate these samples against reference long-run Gibbs draws using a classifier-based Fréchet Inception Distance (FID)~\cite{heusel2017gans} and the class-total variation (class-TV). Empirically, CVSI strictly outperforms both baselines (Fig.~\ref{fig:mnist_results}). It reduces FID by a full order of magnitude compared to TSI, consistently improves upon DSI, and generates visual samples that accurately reflect the target. 

Interestingly, diffusion sampling with CVSI achieves a lower FID than the long-run Gibbs reference itself (dashed floor in Fig.~\ref{fig:mnist_results}). This occurs because the FID classifier was trained on \emph{real} MNIST images, and the block-Gibbs sampler is only \emph{asymptotically} exact, while integrating the score over the diffusion trajectory benefits from the smoothed diffused energy landscape, which improves mode mixing. This is also qualitatively apparent in the visualized samples in Fig.~\ref{fig:mnist_results}a (an extended set of generated samples is provided in Supplementary Fig.~\ref{fig:mnist_samples_full}).
While both DSI and CVSI reproduce the correct mode balance, CVSI's advantage is decisive under strictly limited Monte Carlo budgets due to its lower variance on the MNIST energy model compared with the other estimators (Supplementary Fig.~\ref{fig:mnist_variance}), which precisely mirrors the theoretical insights established in our earlier analytic benchmarks.

%While both DSI and CVSI reproduce the correct mode balance, CVSI's advantage is decisive under strictly limited Monte Carlo budgets (Supplementary Fig.~\ref{fig:mnist_quality_vs_K}). This precisely mirrors the theoretical insights established in earlier analytic benchmarks, as demonstrated by the estimators' empirical variance on the image model (Supplementary Fig.~\ref{fig:mnist_variance}).

%\paragraph{Sampling from Lennard-Jones potentials.}

\section{Discussion}\label{discussion}

Sampling from unnormalized probability densities remains a critical bottleneck across the physical and computational sciences. Although diffusion models offer a compelling generative framework for these tasks, current methods rely on two inherently sub-optimal score estimators: the Denoising Score Identity (DSI) and the Target Score Identity (TSI). Because these estimators exhibit high variance at opposing noise levels, they become computationally expensive and structurally unstable in complex, high-dimensional spaces.

We resolve this dichotomy by introducing the Control Variate Score Identity (CVSI). By framing score estimation within the control variate framework, CVSI encompasses DSI and TSI as limiting cases of a unified, unbiased estimator. Combined with our analytically optimal, state- and time-dependent interpolation coefficient $c^*(\mathbf{x}(t), t)$, CVSI strictly minimizes estimator variance across the entire diffusion trajectory, effectively eliminating the need for brittle heuristic weighting schemes.

Empirically, CVSI demonstrates significant advantages across two primary settings. In data-free sampler learning via importance sampling, CVSI drastically improves sample efficiency and fundamentally stabilizes training dynamics, preventing the mode collapse often seen with a standard TSI estimator. During training-free diffusion sampling, CVSI achieves near-optimal performance utilizing up to six times fewer Monte Carlo posterior samples than baselines. Because each posterior sample requires an (expensive) energy function evaluation (NFE), this drastic reduction in sample complexity makes simulations with (so far) intractable, high-dimensional energy landscapes computationally more feasible. Our experiments using a GB--RBM for small images provide evidence that these efficiency and performance gains can scale to more complex, high-dimensional energy models.
A key insight of our approach is that by stabilizing data-free training, CVSI circumvents the need for massive pre-computed datasets required for training standard approaches, which, for instance, can facilitate the generative exploration of rare events and states in chemical spaces \cite{Liu2023, Park2026}.
In a broader context, our novel framework contributes to the long-lasting quest of alleviating the bias-variance dilemma (cf. Geman \etal~\cite{geman1992neural}). By strictly minimizing the variance while preserving an unbiased score estimate, our work demonstrates that addressing this dilemma is crucial for achieving better performance in sampling and more general unsupervised generative models such as diffusion models.

However, a remaining limitation is the base computational overhead. Estimating the score per training sample $\mathbf{x}(t)$ still requires computing expectations over the posterior $q(\mathbf{x}(0)|\mathbf{x}(t))$. When analytical posteriors are unavailable, this necessitates Markov chain Monte Carlo sampling or importance sampling methods. Although CVSI significantly reduces the total number of posterior samples required, it remains a bottleneck for scaling, say, to ultra-large biochemical systems. 

Future work will explore coupling CVSI with amortized approximations of the posterior to bypass MCMC sampling entirely, further reducing the required NFEs per training step. By bridging the gap between rigorous variance reduction and scalable posterior sampling, we anticipate CVSI will serve as a foundational mechanism for advancing highly efficient sampling across the computational sciences.

\section{Online Methods}\label{methods}

\subsection{Diffusion models}
Consider a (non-zero and differentiable) target probability distribution $p(\mathbf{x})$ on $\mathcal{X} \subseteq \mathbb{R}^d$ defined by an unnormalized energy function $E(\mathbf{x})$, such that $p(\mathbf{x}) \propto \exp(-E(\mathbf{x}))$. The forward diffusion process is defined by the stochastic differential equation (SDE)
\begin{equation}
    \dd \mathbf{x}(t) = f(t) \mathbf{x}(t) \, \dd t + g(t) \, \dd \mathbf{w}(t), \quad \mathbf{x}(0) \sim p(\mathbf{x}),
    \label{eq:methods_fwd_sde}
\end{equation}
for $t \in [0,1]$, where $\mathbf{w}(t)$ denotes the standard multivariate Wiener process. This process produces a perturbation kernel $q(\mathbf{x}(t)|\mathbf{x}(0)) = \mathcal{N}(\mathbf{x}(t); a(t)\mathbf{x}(0), b(t)^2 \mathbf{I})$, where the drift and diffusion coefficients satisfy $f(t) = \dot{a}(t)/a(t)$ and $g^2(t) = 2b(t)/a(t)[a(t)\dot{b}(t) - \dot{a}(t)b(t)]$, respectively. Sampling from the target distribution requires simulating the corresponding reverse-time SDE, initialized at $q_1$:
\begin{equation}
\begin{split}
    \dd \tilde{\mathbf{x}}(t) &= \left[ f(t) \tilde{\mathbf{x}}(t) - \frac{1+ \lambda^2}{2} {g(t)}^2 \nabla_{\tilde{\mathbf{x}}(t)} \log q_t(\tilde{\mathbf{x}}(t))\right] \dd \tilde{t} + \lambda {g(t)} \, \dd \tilde{\mathbf{w}}(t),
\end{split}
\label{eq:methods_rev_sde}
\end{equation}
where $\tilde{t}=1-t$ denotes time reversal \cite{ANDERSON1982313}. While any $\lambda \geq 0$ returns the same marginal distributions $\{q_t(\mathbf{x})\}_{t \in [0,1]}$, setting $\lambda=1$ recovers the standard time-reversal SDE and the limit $\lambda=0$ describes the probability flow ODE \cite{song2021score, diff_NF}. 

Simulating Eq.~\eqref{eq:methods_rev_sde} requires accurately estimating the intractable marginal score $\nabla_{\mathbf{x}(t)} \log q_t(\mathbf{x}(t))$. Standard estimation relies on either the Denoising Score Identity (DSI) or the Target Score Identity (TSI). The DSI leverages available data samples to express the score as an expectation over the conditional perturbation kernel:
\begin{equation}
    \nabla_{\mathbf{x}(t)} \log q_t(\mathbf{x}(t)) = \mathbb{E}_{q(\mathbf{x}(0)|\mathbf{x}(t))}\left[ \nabla_{\mathbf{x}(t)} \log q(\mathbf{x}(t)|\mathbf{x}(0)) \right].
    \label{eq:methods_dsi}
\end{equation}
Alternatively, when only the unnormalized target energy is known, the TSI reformulates the marginal score directly in terms of the target energy gradient:
\begin{equation}
    \nabla_{\mathbf{x}(t)} \log q_t(\mathbf{x}(t)) = \frac{1}{a(t)} \mathbb{E}_{q(\mathbf{x}(0)|\mathbf{x}(t))} \left[ \nabla_{\mathbf{x}(0)} \log p(\mathbf{x}(0)) \right],
    \label{eq:methods_tsi}
\end{equation}
where $q(\mathbf{x}(0)|\mathbf{x}(t))$ is the intractable diffusion posterior (see De Bortoli \etal~\cite{de2024target} and Supplementary Section~\ref{app:proof_proposition1} for the proof).

\subsection{Our control variate score identity (CVSI)}
% To systematically mitigate the variance of the TSI estimator, we use the control variates framework. We propose using the score of the posterior distribution, $h(\mathbf{x}(0), t) = \nabla_{\mathbf{x}(0)} \log q(\mathbf{x}(0)|\mathbf{x}(t))$, as the control variate. By Bayes' theorem, it is highly correlated with the primary TSI integrand $a(t)^{-1} \nabla_{\mathbf{x}(0)} \log p(\mathbf{x}(0))$. Furthermore, standard regularity conditions guarantee that its expectation under the posterior vanishes, i.e. $\mathbb{E}_{q(\mathbf{x}(0)|\mathbf{x}(t))}[h(\mathbf{x}(0), t)] = 0$ (Supplementary Section~\ref{app:proof_proposition1}).  % For a given state- and time-dependent coefficient $c(\mathbf{x}(t), t) \in \mathbb{R}$, we define the Control Variate Score Identity (CVSI) estimator $\hat{\mathbf{s}}_\mathrm{CV}(\mathbf{x}(t))$ as: % \begin{equation} %     \hat{\mathbf{s}}_\mathrm{CV}(\mathbf{x}(t)) = \mathbb{E}_{q(\mathbf{x}(0) | \mathbf{x}(t))} \left[ \frac{1}{a(t)} \nabla_{\mathbf{x}(0)} \log p(\mathbf{x}(0)) - c(\mathbf{x}(t), t) \nabla_{\mathbf{x}(0)} \log q(\mathbf{x}(0) | \mathbf{x}(t)) \right]. %     \label{eq:methods_cvsi} % \end{equation} % By the linearity of expectation, $\hat{\mathbf{s}}_\mathrm{CV}$ is an unbiased estimator of the marginal score $\nabla_{\mathbf{x}(t)} \log q_t(\mathbf{x}(t))$ for any choice of $c(\mathbf{x}(t), t)$ (Supplementary Section~\ref{app:proof_optimal_c}). 

As introduced in the main text, we systematically mitigate the variance of the TSI estimator by employing the posterior score $h(\mathbf{x}(0), t) = \nabla_{\mathbf{x}(0)} \log q(\mathbf{x}(0)|\mathbf{x}(t))$ as a control variate. Standard regularity conditions guarantee that its expectation under the posterior vanishes, $\mathbb{E}_{q(\mathbf{x}(0)|\mathbf{x}(t))}[h(\mathbf{x}(0), t)] = 0$ (Supplementary Section~\ref{app:proof_proposition1}). This yields the unbiased CVSI estimator $\hat{\mathbf{s}}_\mathrm{CV}$ in Eq.~\eqref{eq:cvsi_result}, where by the linearity of expectation, $\hat{\mathbf{s}}_\mathrm{CV}$ is an unbiased estimator of the marginal score $\nabla_{\mathbf{x}(t)} \log q_t(\mathbf{x}(t))$ for any choice of $c(\mathbf{x}(t), t)$ (Supplementary Section~\ref{app:proof_optimal_c}). Given the control variate, we can derive the optimal control coefficient that minimizes the variance.

\subsubsection{Optimal control coefficient}
The variance of the CVSI estimator in Eq.~\eqref{eq:cvsi_result} is governed by the quadratic formulation $\Var(\hat{\mathbf{s}}_\mathrm{CV}) = \Var(g) + c(\mathbf{x}(t), t)^2\Var(h) - 2c(\mathbf{x}(t), t)\Cov(g, h)$, where $g(\mathbf{x}(0), t) = a(t)^{-1} \nabla_{\mathbf{x}(0)} \log p(\mathbf{x}(0))$. Minimizing this variance with respect to $c(\mathbf{x}(t), t)$ produces the optimal coefficient $c^*(\mathbf{x}(t), t) = \Cov(g, h) / \Var(h)$ (Supplementary Section~\ref{app:proof_c}). Because evaluating the posterior score $h(\mathbf{x}(0), t) = \nabla_{\mathbf{x}(0)} \log q(\mathbf{x}(0)|\mathbf{x}(t))$ is computationally intractable in many cases, we reformulate $c^*(\mathbf{x}(t), t)$ entirely in terms of the accessible target score $\mathbf{s}_p = \nabla_{\mathbf{x}(0)} \log p(\mathbf{x}(0))$ and the analytically known perturbation kernel score $\mathbf{s}_{t|0} = \nabla_{\mathbf{x}(t)} \log q(\mathbf{x}(t)|\mathbf{x}(0))$, which gives the tractable optimal coefficient in Eq.~\eqref{eq:opt_c_t_result}. For systems defined by a Boltzmann distribution at temperature $\tau$, $\mathbf{s}_p$ trivially expands to $-\nabla_{\mathbf{x}(0)} E(\mathbf{x}(0)) / \tau$, which allows for the direct computation of optimal variance reduction in physical systems where the energy function is known. Supplementary Section~\ref{app:derivation_cstar} provides more details.

Although the control coefficient $c(\mathbf{x}(t), t)$ could theoretically be extended to a vector-valued function, with an independent control variable for each spatial dimension, our empirical evaluations demonstrated that it provided no significant performance advantage over a spatially aggregated scalar coefficient. Consequently, we use a single scalar function $c(\mathbf{x}(t), t)$ across all spatial dimensions, for a given state $\mathbf{x}(t)$ and time step $t$. The generalized variance and covariance scalars are denoted as $\Var(\mathbf{u}) \coloneqq \mathbb{E}\left[\| \mathbf{u} - \mathbb{E}[\mathbf{u}] \|^2\right] = \text{Tr}(\Sigma_{\mathbf{u}})$ and $\Cov(\mathbf{u}, \mathbf{v}) \coloneqq \mathbb{E}\left[\langle \mathbf{u} - \mathbb{E}[\mathbf{u}], \mathbf{v} -  \mathbb{E}[\mathbf{v}] \rangle\right] = \text{Tr}(\Sigma_{\mathbf{u}\mathbf{v}})$.

\subsubsection{Optimal score interpolation}
While the original definition of the CVSI estimator in Eq.~\eqref{eq:cvsi_result} is a direct result from applying the control variates framework, applying Bayes' rule on the posterior and utilizing the Gaussian symmetry property $\nabla_{\mathbf{x}(0)} \log q(\mathbf{x}(t) | \mathbf{x}(0)) = - a(t) \nabla_{\mathbf{x}(t)} \log q(\mathbf{x}(t) | \mathbf{x}(0))$ allows us to restructure the CVSI estimator into a convex combination of TSI and DSI presented in Eq.~\eqref{eq:interpolated_result} (Supplementary Section~\ref{app:derivation_interpolated}).

\subsection{Data-free sampler learning (iDEM)}
We test CVSI within the Iterated Denoising Energy Matching (iDEM) framework \cite{akhound2024iterated} for data-free sampler training. iDEM relies on Importance Sampling (IS) to approximate the posterior expectation $q(\mathbf{x}(0)|\mathbf{x}(t))$. Given a proposal distribution $\pi(\mathbf{x}(0)|\mathbf{x}(t))$, typically $\mathcal{N}(\mathbf{x}(0); \mathbf{x}(t)/a(t), (b(t)/a(t))^2 \mathbf{I})$, standard iDEM trains a diffusion sampler by minimizing a score matching loss \cite{score_matching, vincent}, where the target is an IS estimator of the high-variance TSI formulation:
\begin{align}
    \nabla_{\mathbf{x}(t)} \log q_t(\mathbf{x}(t)) \approx \frac{1}{a(t)} \frac{1}{K} \sum_{k=1}^K w(\mathbf{x}^{(k)}(0); \mathbf{x}(t)) \, \nabla_{\mathbf{x}(0)}\log p(\mathbf{x}^{(k)}(0)),
\end{align}
where $\mathbf{x}^{(k)}(0) \sim \pi(\mathbf{x}(0)|\mathbf{x}(t))$ and the importance weights are defined as:
\begin{equation}
    w(\mathbf{x}(0); \mathbf{x}(t)) = \frac{q(\mathbf{x}(0)|\mathbf{x}(t))}{\pi(\mathbf{x}(0)|\mathbf{x}(t))} = \frac{p(\mathbf{x}(0))}{a(t)^D \, q_t(\mathbf{x}(t))} \,.
\end{equation}
See Supplementary Section~\ref{app:idem} for details.
Since this formulation relies directly on the TSI integrand $\nabla_{\mathbf{x}(0)} \log p(\mathbf{x}(0))$, it inherits its high variance at large noise levels, which is further exacerbated by the variance of the importance weights, increasing the sample complexity. Therefore, we test our CVSI by replacing the standard TSI integrand with our interpolated formulation (Equation~\ref{eq:interpolated_result}). For a finite batch of $K$ importance samples $\mathbf{x}^{(k)}(0) \sim \pi(\mathbf{x}(0)|\mathbf{x}(t))$, the target CVSI-iDEM estimator is defined as:
\begin{equation}
\begin{split}
    \hat{\mathbf{s}}_{\mathrm{CV-iDEM}} &= \sum_{k=1}^K w_k \bigg[ \frac{1-\hat{c}^*(\mathbf{x}(t), t)}{a(t)} \nabla_{\mathbf{x}(0)} \log p(\mathbf{x}^{(k)}(0)) \\
    &\quad + \hat{c}^*(\mathbf{x}(t), t) \nabla_{\mathbf{x}(t)} \log q(\mathbf{x}(t)|\mathbf{x}^{(k)}(0)) \bigg].
\end{split}
\label{eq:methods_idem}
\end{equation}
The normalized importance weights are implemented via a softmax operation, $w_k = p(\mathbf{x}^{(k)}(0)) / \sum_{j=1}^K p(\mathbf{x}^{(j)}(0))$, to ensure numerical stability, which is equivalent to the theoretical importance weights $w(\mathbf{x}^{(k)}(0); \mathbf{x}(t)) \propto p(\mathbf{x}^{(k)}(0)) / q_t(\mathbf{x}(t))$ since the normalization constant $q_t(\mathbf{x}(t))$ and the factor $a(t)^D$ cancel in the softmax operation. The interpolation weight $\hat{{c}}^*(\mathbf{x}(t), t)$ is evaluated empirically at each step for each sample $\mathbf{x}(t)$ using the unweighted sample variance and covariance of the proposal samples. While reweighting these statistics by $w_k$ would theoretically target the true posterior variance, we found that using the unweighted statistics results in more stable training dynamics and better results overall. We note that, as with standard iDEM, the use of Self-Normalized Importance Sampling (SNIS) introduces a bias of order $\mathcal{O}(1/K)$ to the score estimate.
Furthermore, estimating the control coefficient $\hat{c}^*(t)$ on the same set of samples used to compute the score introduces an additional bias.

\subsection{Mitigating error-induced temperature inflation}
During data-free learning, scaling target distributions to higher dimensions increases the score approximation errors, particularly during early training epochs, where the replay buffer has many noise samples. We model this error as a time-dependent Gaussian process, $\boldsymbol{\xi}(t) \sim \mathcal{N}\left(\mathbf{0}, \sigma_{\text{err}}^2(t) \mathbf{I}\right)$, such that the approximated score is $\mathbf{s}_\theta(\mathbf{x}(t), t) = \nabla_{\mathbf{x}(t)} \log q_t(\mathbf{x}(t)) + \boldsymbol{\xi}(t)$. When this score is substituted into the discretized version of the reverse SDE (Eq.~\eqref{eq:methods_rev_sde}) with step size $\Delta t$, the error term acts as an additional perturbation. This artificially grows the effective variance of the trajectory update:
\begin{equation}
\begin{split}
    \Var(\Delta \tilde{\mathbf{x}}_t | \tilde{\mathbf{x}}(t)) &\approx \lambda^2 g(t)^2 \Delta t + \left(\frac{1+\lambda^2}{2}\right)^2 g(t)^4 \sigma_{\text{err}}^2(t) (\Delta t)^2,
\end{split}
\end{equation}
which acts as an increase in the sampling temperature, leading to overly noisy samples that populate the replay buffer and create a deteriorating feedback loop. While in the theoretical continuous limit $\Delta t \to 0$ the error contribution to the variance vanishes, in the discrete case it increases with larger step sizes. To mitigate this temperature inflation and maintain the target variance dictated by $\lambda^2$, we decouple the explicit noise injected by the Wiener process from the drift formulation, and we introduce a new effective diffusion scale:
\begin{equation}
    \lambda_{\text{eff}}(t, \Delta t) = \sqrt{\lambda^2 - \left( \frac{1+\lambda^2}{2} \right)^2 g(t)^2 \sigma_{\text{err}}^2(t) \Delta t}.
    \label{eq:lambda_eff}
\end{equation}
By replacing $\lambda$ with $\lambda_\text{eff} \leq \lambda$ solely in the diffusion term (leaving the drift and $\lambda$ unchanged), we safely counteract the stochasticity induced by the error (Supplementary Section~\ref{app:gaussian_error_analysis}). In practice, manually setting a constant $\lambda_\text{eff}$ is sufficient to break the feedback loop and stabilize the training. We use $\lambda_\text{eff}=0.5$ for the high-dimensional GMM case, and the default $\lambda_\text{eff}=1.0$ for the DW-4 and 2D GMM case.

\subsection{Exact posterior sampling for the Gaussian--Bernoulli RBM}
\label{sec:gbrbm_exact_sampling}

To evaluate CVSI on a high-dimensional image energy landscape we use a Gaussian--Bernoulli restricted Boltzmann machine (GB--RBM)~\cite{welling2004exponential}, and take advantage of its bipartite structure to tractably sample from the diffusion posterior. We identify the target variable $\mathbf{x}(0)$ with the continuous visible units $\mathbf{v}\in\mathbb{R}^D$ of an RBM, i.e. $\mathbf{x}(0) \equiv \mathbf{v}$, coupled to $M$ binary latent units $\mathbf{h}\in\{0,1\}^M$. The joint probability is defined as $p(\mathbf{v},\mathbf{h})\propto\exp(-E(\mathbf{v},\mathbf{h}))$ with joint energy:
\begin{equation}
    E(\mathbf{v},\mathbf{h}) = \frac{\lVert \mathbf{v}-\bv\rVert^2}{2\sigma^2} - \frac{\mathbf{h}^\top W \mathbf{v}}{\sigma^2} - \bh^\top \mathbf{h},
    \label{eq:gbrbm_energy_methods}
\end{equation}
with weight matrix $W\in\mathbb{R}^{M\times D}$ and visible noise scale $\sigma>0$. The vectors $\mathbf{d}_v\in\mathbb{R}^D$ and $\mathbf{d}_h\in\mathbb{R}^M$ are the visible and hidden model biases, respectively. Once the model parameters ($W$, $\mathbf{d}_v$, and $\mathbf{d}_h$) are learned, e.g. via contrastive divergence, the energy model is fixed.
For a fixed diffused state $\mathbf{x}(t)$, the exact diffusion posterior $q(\mathbf{x}(0)|\mathbf{x}(t)) \equiv q(\mathbf{v}|\mathbf{x}(t)) \propto p(\mathbf{v}) \mathcal{N}(\mathbf{x}(t); a(t)\mathbf{v}, b(t)^2\mathbf{I})$ is intractable, because the marginal distribution $p(\mathbf{v}) = \sum_{\mathbf{h}} p(\mathbf{v},\mathbf{h})$ has an intractable normalizing constant that prevents direct sampling. To overcome this, we extend the standard bipartite structure of the GB-RBM directly into the diffusion framework. By reintroducing the latent units $\mathbf{h}$, we define the augmented diffusion posterior $q(\mathbf{v},\mathbf{h}|\mathbf{x}(t)) \propto \exp(-E(\mathbf{v},\mathbf{h})) \mathcal{N}(\mathbf{x}(t); a(t)\mathbf{v}, b(t)^2\mathbf{I})$. As formally proven in Supplementary Section~\ref{app:gbrbm_derivations}, this augmented posterior admits closed-form full conditionals:
\begin{align}
    q(\mathbf{h}|\mathbf{v},\mathbf{x}(t)) &= \prod_{j=1}^{M} \mathrm{Bernoulli}\!\left(h_j;\, \sigm\!\left(d_{h,j} + \frac{(W\mathbf{v})_j}{\sigma^2}\right)\right), \label{eq:h_given_v_tethered} \\
    q(\mathbf{v}|\mathbf{h},\mathbf{x}(t)) &= \mathcal{N}\big(\mathbf{v};\, \boldsymbol{\mu}(\mathbf{h},\mathbf{x}(t)),\, \Lambda(t)^{-1}\mathbf{I}\big), \label{eq:v_given_h_tethered}
\end{align}
where $\sigm(\cdot)$ is the logistic sigmoid, $d_{h,j}$ is the $j$-th entry of the offset $\mathbf{d}_h$, the precision is $\Lambda(t) = 1 / \sigma^{2} + \left(a(t) / b(t)\right)^{2}$, and the mean is $\boldsymbol{\mu}(\mathbf{h},\mathbf{x}(t)) = \Lambda(t)^{-1} \big( (\mathbf{d}_v+W^\top\mathbf{h})/\sigma^2 + a(t)\mathbf{x}(t)/b(t)^2 \big)$.

Given the fixed energy model, this augmented formulation allows an exact block-Gibbs sampler for the diffusion posterior from any state $\mathbf{x}(t)$. Specifically, starting from $\mathbf{v}^{(0)}\sim\mathcal{N}(\mathbf{x}(t)/a(t),\,(b(t)/a(t))^2\mathbf{I})$, we alternate draws from Eqs.~\eqref{eq:h_given_v_tethered} and \eqref{eq:v_given_h_tethered}. Because both conditionals have full support, standard Gibbs sampling theory \cite{geman1984stochastic} guarantees the chain's unique stationary distribution is exactly $q(\mathbf{v},\mathbf{h}|\mathbf{x}(t))$. Furthermore, because the posterior precision $\Lambda(t)$ grows without bound as $t\to 0$, the observation $\mathbf{x}(t)$ pins the chain to a more local neighborhood. Mixing is consequently local rather than global, requiring only a modest number of iterations per chain. By running $K$ independent chains in parallel we obtain asymptotically unbiased draws $\{\mathbf{v}^{(k)}\}_{k=1}^K\sim q(\mathbf{v}|\mathbf{x}(t))$.

These draws can be used directly in the TSI, DSI, and CVSI Monte Carlo estimators without any importance-weighting terms or Metropolis corrections:
\begin{align}
    \widehat{\mathrm{TSI}} = -\frac{1}{K a(t)}\sum_{k=1}^K \nabla_{\mathbf{v}^{(k)}} F(\mathbf{v}^{(k)}), \quad
    \widehat{\mathrm{DSI}} = \frac{1}{K}\sum_{k=1}^{K} \frac{a(t)\mathbf{v}^{(k)} - \mathbf{x}(t)}{b(t)^2}. \label{eq:dsi_mc}
\end{align}
While $p(\mathbf{v})$ is intractable, its unnormalized free energy $F(\mathbf{v})$ is obtained in closed form (see Supplementary Section~\ref{app:gbrbm_derivations}). Consequently, the TSI integrand $\nabla_{\mathbf{v}}\log p(\mathbf{v}) = -\nabla_{\mathbf{v}} F(\mathbf{v})$ is available for any sample. The CVSI interpolation weight $\hat{c}^*(\mathbf{x}(t),t)$ is likewise evaluated directly from the unweighted sample variance and covariance of the posterior draws.

\subsection{Experimental configuration}
We test our method on training-free diffusion sampling and data-free sampler learning. For training-free sampling, analytical evaluations were performed using $d$-dimensional Gaussian Mixture Models (GMMs). To ensure the difficulty of the problem scales appropriately with dimensionality, we generate the mixture parameters randomly. Specifically, the mixture weights $\pi^i$ are drawn uniformly and normalized, and the means are sampled from $\boldsymbol{\mu}^i \sim \mathcal{N}(\mathbf{0}, s^2 d \cdot \mathbf{I})$ with scale $s=10$, ensuring that the modes remain distinguishable as dimension $d$ increases. The covariance matrices are sampled from a Wishart distribution $\boldsymbol{\Sigma}^i \sim \mathcal{W}(\nu, \mathbf{I})$ with degrees of freedom $\nu = 2d$. The diffused marginal and posterior distributions of the GMM admit exact closed-form analytical solutions, which are stated in Supplementary Section~\ref{app:analytical_gmm}. 

For data-free sampler learning within the iDEM framework, multi-modal evaluations were conducted on both complex GMMs and the 4-particle Double-Well (DW-4) physical system. The DW-4 energy landscape is parameterized by pairwise distances $d_{ij} = ||\mathbf{x}_i - \mathbf{x}_j||_2$ with constants $b=-4$, $c=0.9$, $d_0=4$, and temperature $\tau=1$, defining a distribution invariant to rotations and permutations. Unless otherwise stated, we use the same models, noise schedules and hyperparameters as in iDEM~\cite{akhound2024iterated}, where an MLP is used for the GMM case and an EGNN for the DW-4 case.

For high-dimensional image evaluation, we fit a Gaussian--Bernoulli RBM ($D=196$, $M=124$) on $50{,}000$ $14\times14$ MNIST training images. To prevent mode collapse during training, we utilize persistent contrastive divergence (PCD-1)~\cite{tieleman2008training} with a single Gibbs step per update, maintaining a continuous pool of 256 particles initialized from the training data. Images were pre-standardized to zero mean and unit variance, allowing us to fix the visible noise scale at $\sigma=1$. The model was trained for 200 epochs using the Adam optimizer (learning rate $10^{-4}$, weight decay $10^{-4}$, batch size 256, gradient clipping norm 10). Negative-phase visible units were sampled stochastically from their conditional Gaussian, $\mathbf{v}\sim\mathcal{N}(\mathbf{d}_v+W^\top\mathbf{h},\,\sigma^2\mathbf{I})$, rather than using their conditional mean. 
For sampling with the different estimators, we simulate the reverse SDE (Eq.~\eqref{eq:methods_rev_sde}) under a VE schedule ($a(t)=1$, $b(t)>0$) throughout. Sample quality was evaluated using a classifier-FID \cite{heusel2017gans}, computed on the feature space of a pre-trained MNIST classifier, and the class-total variation (class-TV) between the predicted-label histograms of generated and reference samples.

\subsection*{Acknowledgements}
K.R.M.\ and K.K.\ were partly funded by the German Ministry for Education and Research (BMBF) as BIFOLD – Berlin Institute for the Foundations of Learning and Data - under Grants 01IS14013A-E, 01GQ1115, 01GQ0850, 01IS18025A, 031L0207D, and 01IS18037A and DFG.
K.R.M.\ was also supported by the Institute of Information \& communications Technology Planning \& Evaluation (IITP) grants funded by the Korea government (MSIT) (No.\ RS-2019-II190079, Artificial Intelligence Graduate School Program of Korea University and No.\ RS-2024-00457882, AI Research Hub Project), and also by DFG and HFA. KRM  was furthermore supported in parts by the Korea University Grant. Correspondence to KRM, OTU and AD. 

\section*{Data availability}
The MNIST dataset analyzed during the current study is publicly available. The data used for the evaluations will be made public on GitHub at https://github.com/khaledkah/cvsi.

\section*{Code availability}
The source code used to implement the Control Variate Score Identity (CVSI) framework and generate the results reported in this study will be made available upon publication in the open-source repository at https://github.com/khaledkah/cvsi.

\backmatter

\bibliography{main}

@article{geman1992neural,
  title={Neural networks and the bias/variance dilemma},
  author={Geman, Stuart and Bienenstock, Elie and Doursat, Ren{\'e}},
  journal={Neural computation},
  volume={4},
  number={1},
  pages={1--58},
  year={1992},
  publisher={MIT Press}
}

@inproceedings{
    adjointsampling,
    title={Adjoint Sampling: Highly Scalable Diffusion Samplers via Adjoint Matching},
    author={Aaron J Havens and Benjamin Kurt Miller and Bing Yan and Carles Domingo-Enrich and Anuroop Sriram and Daniel S. Levine and Brandon M Wood and Bin Hu and Brandon Amos and Brian Karrer and Xiang Fu and Guan-Horng Liu and Ricky T. Q. Chen},
    booktitle={Forty-second International Conference on Machine Learning},
    year={2025},
    url={https://openreview.net/forum?id=6Eg1OrHmg2}
}

@misc{liu2025adjointschrodingerbridgesampler,
      title={Adjoint Schr\"odinger Bridge Sampler}, 
      author={Guan-Horng Liu and Jaemoo Choi and Yongxin Chen and Benjamin Kurt Miller and Ricky T. Q. Chen},
      year={2025},
      eprint={2506.22565},
      archivePrefix={arXiv},
      primaryClass={stat.ML},
      url={https://arxiv.org/abs/2506.22565}, 
}

@misc{choi2025noneqsampler,
      title={Non-equilibrium Annealed Adjoint Sampler}, 
      author={Jaemoo Choi and Yongxin Chen and Molei Tao and Guan-Horng Liu},
      year={2025},
      eprint={2506.18165},
      archivePrefix={arXiv},
      primaryClass={cs.LG},
      url={https://arxiv.org/abs/2506.18165}, 
}

@misc{akhoundsadegh2025progressive,
      title={Progressive Inference-Time Annealing of Diffusion Models for Sampling from Boltzmann Densities}, 
      author={Tara Akhound-Sadegh and Jungyoon Lee and Avishek Joey Bose and Valentin De Bortoli and Arnaud Doucet and Michael M. Bronstein and Dominique Beaini and Siamak Ravanbakhsh and Kirill Neklyudov and Alexander Tong},
      year={2025},
      eprint={2506.16471},
      archivePrefix={arXiv},
      primaryClass={cs.LG},
      url={https://arxiv.org/abs/2506.16471}, 
}

@inproceedings{
    zhang2021path,
    title={Path Integral Sampler: A Stochastic Control Approach For Sampling},
    author={Qinsheng Zhang and Yongxin Chen},
    booktitle={International Conference on Learning Representations},
    year={2022},
    url={https://openreview.net/forum?id=_uCb2ynRu7Y}
}

@inproceedings{
    huang2023reverse,
    title={Reverse Diffusion Monte Carlo},
    author={Xunpeng Huang and Hanze Dong and Yifan HAO and Yian Ma and Tong Zhang},
    booktitle={The Twelfth International Conference on Learning Representations},
    year={2024},
    url={https://openreview.net/forum?id=kIPEyMSdFV}
}

@inproceedings{ko2025latent,
  title={Latent Target Score Matching, with an application to Simulation-Based Inference},
  author={Ko, Joohwan and Geffner, Tomas},
  booktitle={Machine Learning and the Physical Sciences Workshop, NeurIPS},
  year={2025}
}

@article{nicoli2020asymptotically,
  title={Asymptotically unbiased estimation of physical observables with neural samplers},
  author={Nicoli, Kim A and Nakajima, Shinichi and Strodthoff, Nils and Samek, Wojciech and M{\"u}ller, Klaus-Robert and Kessel, Pan},
  journal={Physical Review E},
  volume={101},
  number={2},
  pages={023304},
  year={2020},
  publisher={APS}
}

@article{midgley2022flow,
  title={Flow annealed importance sampling bootstrap},
  author={Midgley, Laurence Illing and Stimper, Vincent and Simm, Gregor NC and Sch{\"o}lkopf, Bernhard and Hern{\'a}ndez-Lobato, Jos{\'e} Miguel},
  journal={arXiv preprint arXiv:2208.01893},
  year={2022}
}

@inproceedings{akhound2024iterated,
  title={Iterated denoising energy matching for sampling from {B}oltzmann densities},
  author={Akhound-Sadegh, Tara and Rector-Brooks, Jarrid and Bose, Avishek Joey and Mittal, Sarthak and Lemos, Pablo and Liu, Cheng-Hao and Sendera, Marcin and Ravanbakhsh, Siamak and Gidel, Gauthier and Bengio, Yoshua and Malkin, Nikolay and Tong, Alexander},
  booktitle={International Conference on Machine Learning},
  year={2024}
}

@article{
    berner2024an,
    title={An optimal control perspective on diffusion-based generative modeling},
    author={Julius Berner and Lorenz Richter and Karen Ullrich},
    journal={Transactions on Machine Learning Research},
    issn={2835-8856},
    year={2024},
    url={https://openreview.net/forum?id=oYIjw37pTP},
    note={}
}

@inproceedings{
    vargas2023denoising,
    title={Denoising Diffusion Samplers},
    author={Francisco Vargas and Will Sussman Grathwohl and Arnaud Doucet},
    booktitle={The Eleventh International Conference on Learning Representations },
    year={2023},
    url={https://openreview.net/forum?id=8pvnfTAbu1f}
}

@article{de2024target,
  title={Target score matching},
  author={De Bortoli, Valentin and Hutchinson, Michael and Wirnsberger, Peter and Doucet, Arnaud},
  journal={arXiv preprint arXiv:2402.08667},
  year={2024}
}

@inproceedings{DDPM_Ho,
  author    = {Ho, Jonathan and Jain, Ajay and Abbeel, Pieter},
  booktitle = {Advances in Neural Information Processing Systems},
  title     = {Denoising Diffusion Probabilistic Models},
  year      = {2020},
  editor    = {H. Larochelle and M. Ranzato and R. Hadsell and M.F. Balcan and H. Lin},
  pages     = {6840--6851},
  publisher = {Curran Associates, Inc.},
  volume    = {33},
  url       = {https://proceedings.neurips.cc/paper_files/paper/2020/file/4c5bcfec8584af0d967f1ab10179ca4b-Paper.pdf}
}

@article{kahouli2025tv_snr,
  title={Disentangling Total-Variance and Signal-to-Noise-Ratio Improves Diffusion Models},
  author={Kahouli, Khaled and Ripken, Winfried and Gugler, Stefan and Unke, Oliver T and M{\"u}ller, Klaus-Robert and Nakajima, Shinichi},
  journal={arXiv preprint arXiv:2502.08598},
  year={2025}
}

@article{lemieux2014control,
  title={Control variates},
  author={Lemieux, Christiane},
  journal={Wiley StatsRef: Statistics Reference Online},
  pages={1--8},
  year={2014},
  publisher={Wiley Online Library}
}

@inproceedings{diff_mod_sohl,
  author    = {Sohl-Dickstein, Jascha and Weiss, Eric and Maheswaranathan, Niru and Ganguli, Surya},
  booktitle = {Proceedings of the 32nd International Conference on Machine Learning},
  title     = {Deep Unsupervised Learning using Nonequilibrium Thermodynamics},
  year      = {2015},
  address   = {Lille, France},
  editor    = {Bach, Francis and Blei, David},
  month     = {07--09 Jul},
  pages     = {2256--2265},
  publisher = {PMLR},
  series    = {Proceedings of Machine Learning Research},
  volume    = {37},
  url       = {https://proceedings.mlr.press/v37/sohl-dickstein15.html}
}

@article{
    bioemu,
    author = {Sarah Lewis  and Tim Hempel  and José Jiménez-Luna  and Michael Gastegger  and Yu Xie  and Andrew Y. K. Foong  and Victor García Satorras  and Osama Abdin  and Bastiaan S. Veeling  and Iryna Zaporozhets  and Yaoyi Chen  and Soojung Yang  and Adam E. Foster  and Arne Schneuing  and Jigyasa Nigam  and Federico Barbero  and Vincent Stimper  and Andrew Campbell  and Jason Yim  and Marten Lienen  and Yu Shi  and Shuxin Zheng  and Hannes Schulz  and Usman Munir  and Roberto Sordillo  and Ryota Tomioka  and Cecilia Clementi  and Frank Noé },
    title = {Scalable emulation of protein equilibrium ensembles with generative deep learning},
    journal = {Science},
    volume = {389},
    number = {6761},
    pages = {eadv9817},
    year = {2025},
    doi = {10.1126/science.adv9817},
    URL = {https://www.science.org/doi/abs/10.1126/science.adv9817},
    eprint = {https://www.science.org/doi/pdf/10.1126/science.adv9817},
}

@article{vincent,
  author  = {Vincent, Pascal},
  journal = {Neural Computation},
  title   = {A Connection Between Score Matching and Denoising Autoencoders},
  year    = {2011},
  number  = {7},
  pages   = {1661-1674},
  volume  = {23},
  doi     = {10.1162/NECO_a_00142}
}

@inproceedings{song2021score,
  author    = {Yang Song and Jascha Sohl-Dickstein and Diederik P Kingma and Abhishek Kumar and Stefano Ermon and Ben Poole},
  booktitle = {International Conference on Learning Representations},
  title     = {Score-Based Generative Modeling through Stochastic Differential Equations},
  year      = {2021},
  url       = {https://openreview.net/forum?id=PxTIG12RRHS},
  pages = {37799--37812}
}

@article{score_matching,
  author  = {Aapo Hyv{{\"a}}rinen},
  journal = {Journal of Machine Learning Research},
  title   = {Estimation of Non-Normalized Statistical Models by Score Matching},
  year    = {2005},
  number  = {24},
  pages   = {695--709},
  volume  = {6},
  url     = {http://jmlr.org/papers/v6/hyvarinen05a.html}
}

@book{DoucetFG01,
  editor       = {Arnaud Doucet and
                  Nando de Freitas and
                  Neil J. Gordon},
  title        = {Sequential Monte Carlo Methods in Practice},
  series       = {Statistics for Engineering and Information Science},
  publisher    = {Springer},
  year         = {2001},
  url          = {https://doi.org/10.1007/978-1-4757-3437-9},
  doi          = {10.1007/978-1-4757-3437-9},
  isbn         = {978-1-4419-2887-0},
  bibsource    = {dblp computer science bibliography, https://dblp.org}
}

@Article{Andrieu2003,
    author={Andrieu, Christophe
    and de Freitas, Nando
    and Doucet, Arnaud
    and Jordan, Michael I.},
    title={An Introduction to MCMC for Machine Learning},
    journal={Machine Learning},
    year={2003},
    month={Jan},
    day={01},
    volume={50},
    number={1},
    pages={5-43},
    issn={1573-0565},
    doi={10.1023/A:1020281327116},
    url={https://doi.org/10.1023/A:1020281327116}
}

@incollection{neal2011mcmc,
  title={MCMC using Hamiltonian dynamics},
  author={Neal, Radford M.},
  booktitle={Handbook of Markov Chain Monte Carlo},
  volume={54},
  pages={113--162},
  year={2011},
  editor={Brooks, Steve and Gelman, Andrew and Jones, Galin and Meng, Xiao-Li},
  publisher={CRC Press},
}

@article{Roberts_mcmc,
author = {G. O. Roberts and A. Gelman and W. R. Gilks},
title = {{Weak convergence and optimal scaling of random walk Metropolis algorithms}},
volume = {7},
journal = {The Annals of Applied Probability},
number = {1},
publisher = {Institute of Mathematical Statistics},
pages = {110 -- 120},
year = {1997},
doi = {10.1214/aoap/1034625254},
URL = {https://doi.org/10.1214/aoap/1034625254}
}

@Article{Qiao2024,
    author={Qiao, Zhuoran
    and Nie, Weili
    and Vahdat, Arash
    and Miller, Thomas F.
    and Anandkumar, Animashree},
    title={State-specific protein--ligand complex structure prediction with a multiscale deep generative model},
    journal={Nature Machine Intelligence},
    year={2024},
    month={Feb},
    day={01},
    volume={6},
    number={2},
    pages={195-208},
    issn={2522-5839},
    doi={10.1038/s42256-024-00792-z},
    url={https://doi.org/10.1038/s42256-024-00792-z}
}

@Article{Liu2023,
    author={Liu, Shengchao
    and Nie, Weili
    and Wang, Chengpeng
    and Lu, Jiarui
    and Qiao, Zhuoran
    and Liu, Ling
    and Tang, Jian
    and Xiao, Chaowei
    and Anandkumar, Animashree},
    title={Multi-modal molecule structure--text model for text-based retrieval and editing},
    journal={Nature Machine Intelligence},
    year={2023},
    month={Dec},
    day={01},
    volume={5},
    number={12},
    pages={1447-1457},
    issn={2522-5839},
    doi={10.1038/s42256-023-00759-6},
    url={https://doi.org/10.1038/s42256-023-00759-6}
}

@Article{Park2026,
    author={Park, Hyunsoo
    and Walsh, Aron},
    title={Guiding generative models to uncover diverse and novel crystals via reinforcement learning},
    journal={Nature Machine Intelligence},
    year={2026},
    month={Jul},
    day={01},
    volume={8},
    number={7},
    pages={1087-1099},
    issn={2522-5839},
    doi={10.1038/s42256-026-01262-4},
    url={https://doi.org/10.1038/s42256-026-01262-4}
}

@article{CotterMCMC,
 ISSN = {08834237, 21688745},
 URL = {http://www.jstor.org/stable/43288425},
 author = {S. L. Cotter and G. O. Roberts and A. M. Stuart and D. White},
 journal = {Statistical Science},
 number = {3},
 pages = {424--446},
 publisher = {Institute of Mathematical Statistics},
 title = {MCMC Methods for Functions: Modifying Old Algorithms to Make Them Faster},
 urldate = {2026-08-19},
 volume = {28},
 year = {2013}
}

@inproceedings{
    sendera2024improved,
    title={Improved off-policy training of diffusion samplers},
    author={Marcin Sendera and Minsu Kim and Sarthak Mittal and Pablo Lemos and Luca Scimeca and Jarrid Rector-Brooks and Alexandre Adam and Yoshua Bengio and Nikolay Malkin},
    booktitle={The Thirty-eighth Annual Conference on Neural Information Processing Systems},
    year={2024},
    url={https://openreview.net/forum?id=vieIamY2Gi}
    }

@article{ANDERSON1982313,
  author   = {Brian D.O. Anderson},
  journal  = {Stochastic Processes and their Applications},
  title    = {Reverse-time diffusion equation models},
  year     = {1982},
  issn     = {0304-4149},
  number   = {3},
  pages    = {313-326},
  volume   = {12},
  doi      = {https://doi.org/10.1016/0304-4149(82)90051-5},
  url      = {https://www.sciencedirect.com/science/article/pii/0304414982900515}
}

@article{noe2019boltzmann,
  author    = {No{\'e}, Frank and Olsson, Simon and K{\"o}hler, Jonas and Wu, Hao},
  journal   = {Science},
  title     = {Boltzmann generators: Sampling equilibrium states of many-body systems with deep learning},
  year      = {2019},
  number    = {6457},
  pages     = {eaaw1147},
  volume    = {365},
  doi       = {10.1126/science.aaw1147},
  publisher = {American Association for the Advancement of Science}
}

@article{albergo2019flow,
  title={Flow-based generative models for Markov chain Monte Carlo in lattice field theory},
  author={Albergo, Michael S and Kanwar, Gurtej and Shanahan, Phiala E},
  journal={Physical Review D},
  volume={100},
  number={3},
  pages={034515},
  year={2019},
  publisher={APS}
}

@inproceedings{diff_edm,
  author    = {Tero Karras and Miika Aittala and Timo Aila and Samuli Laine},
  booktitle = {Advances in Neural Information Processing Systems},
  title     = {Elucidating the Design Space of Diffusion-Based Generative Models},
  year      = {2022},
  editor    = {Alice H. Oh and Alekh Agarwal and Danielle Belgrave and Kyunghyun Cho},
  url       = {https://openreview.net/forum?id=k7FuTOWMOc7}
}

@inproceedings{diff_NF,
  title     = {Diffusion Normalizing Flow},
  author    = {Qinsheng Zhang and Yongxin Chen},
  booktitle = {Advances in Neural Information Processing Systems},
  editor    = {A. Beygelzimer and Y. Dauphin and P. Liang and J. Wortman Vaughan},
  year      = {2021},
  url       = {https://openreview.net/forum?id=x1Lp2bOlVIo}
}

@inproceedings{grenioux2024stochastic,
  title={Stochastic Localization via Iterative Posterior Sampling},
  author={Grenioux, Louis and Noble, Maxence and Gabri{\'e}, Marylou and Durmus, Alain Oliviero},
  booktitle={International Conference on Machine Learning},
  pages={16337--16376},
  year={2024},
  organization={PMLR}
}

@inproceedings{phillips2024particle,
  title={Particle denoising diffusion sampler},
  author={Phillips, Angus and Dau, Hai-Dang and Hutchinson, Michael John and De Bortoli, Valentin and Deligiannidis, George and Doucet, Arnaud},
  booktitle={Proceedings of the 41st International Conference on Machine Learning},
  year={2024}
}

@inproceedings{welling2004exponential,
 author = {Welling, Max and Rosen-zvi, Michal and Hinton, Geoffrey E},
 booktitle = {Advances in Neural Information Processing Systems},
 editor = {L. Saul and Y. Weiss and L. Bottou},
 pages = {},
 publisher = {MIT Press},
 title = {Exponential Family Harmoniums with an Application to Information Retrieval},
 url = {https://proceedings.neurips.cc/paper_files/paper/2004/file/0e900ad84f63618452210ab8baae0218-Paper.pdf},
 volume = {17},
 year = {2004}
}

@inproceedings{tieleman2008training,
  title={Training restricted Boltzmann machines using approximations to the likelihood gradient},
  author={Tieleman, Tijmen},
  booktitle={Proceedings of the 25th International Conference on Machine Learning},
  pages={1064--1071},
  year={2008}
}

@Article{Jumper2021,
author={Jumper, John
and Evans, Richard
and Pritzel, Alexander
and Green, Tim
and Figurnov, Michael
and Ronneberger, Olaf
and Tunyasuvunakool, Kathryn
and Bates, Russ
and {\v{Z}}{\'i}dek, Augustin
and Potapenko, Anna
and Bridgland, Alex
and Meyer, Clemens
and Kohl, Simon A. A.
and Ballard, Andrew J.
and Cowie, Andrew
and Romera-Paredes, Bernardino
and Nikolov, Stanislav
and Jain, Rishub
and Adler, Jonas
and Back, Trevor
and Petersen, Stig
and Reiman, David
and Clancy, Ellen
and Zielinski, Michal
and Steinegger, Martin
and Pacholska, Michalina
and Berghammer, Tamas
and Bodenstein, Sebastian
and Silver, David
and Vinyals, Oriol
and Senior, Andrew W.
and Kavukcuoglu, Koray
and Kohli, Pushmeet
and Hassabis, Demis},
title={Highly accurate protein structure prediction with AlphaFold},
journal={Nature},
year={2021},
month={Aug},
day={01},
volume={596},
number={7873},
pages={583-589},
issn={1476-4687},
doi={10.1038/s41586-021-03819-2},
url={https://doi.org/10.1038/s41586-021-03819-2}
}

@ARTICLE{geman1984stochastic,
  author={Geman, Stuart and Geman, Donald},
  journal={IEEE Transactions on Pattern Analysis and Machine Intelligence}, 
  title={Stochastic Relaxation, Gibbs Distributions, and the Bayesian Restoration of Images}, 
  year={1984},
  volume={PAMI-6},
  number={6},
  pages={721-741},
  doi={10.1109/TPAMI.1984.4767596}}

@inproceedings{heusel2017gans,
  title={{GAN}s trained by a two time-scale update rule converge to a local {N}ash equilibrium},
  author={Heusel, Martin and Ramsauer, Hubert and Unterthiner, Thomas and Nessler, Bernhard and Hochreiter, Sepp},
  booktitle={Advances in Neural Information Processing Systems},
  volume={30},
  year={2017}
}

\newpage

\section*{Supplementary Information}

\begin{appendices}
\renewcommand{\thefigure}{\arabic{figure}}
\renewcommand{\thetable}{\arabic{table}}
\renewcommand{\theHfigure}{S\arabic{figure}} % Fixes hyperref jump issue
\renewcommand{\theHtable}{S\arabic{table}}   % Fixes hyperref jump issue
\setcounter{figure}{0}
\setcounter{table}{0}

\section{Proof of Eq~\eqref{eq:methods_tsi}}
\label{app:proof_proposition1}
In the following, we aim to prove that:
\begin{equation}
    \nabla_{\mathbf{x}} \log q_t(\mathbf{x}) = \frac{1}{a(t)} \mathbb{E}_{q(\mathbf{x}(0) | \mathbf{x}(t))} \left[ \nabla_{\mathbf{x}} \log p(\mathbf{x}) \right].
\end{equation}
The proof can be performed using either Bayes' rule, see Supplementary Section~\ref{app:proof_proposition1_A}, or integration by parts, see Supplementary Section~\ref{app:proof_proposition1_B}.

\subsection{Derivation using Bayes' rule}
\label{app:proof_proposition1_A}

Starting with the DSI identity:
\begin{align}
    \nabla_{\mathbf{x}} \log q_t(\mathbf{x}) & \overset{\text{DSI}}{=} \mathbb{E}_{q(\mathbf{x}(0)|\mathbf{x}(t))}[\nabla_{\mathbf{x}(t)} \log q(\mathbf{x}(t)|\mathbf{x}(0))] \\
                                             & \overset{\text{Gauss. sym.}}{=} \mathbb{E}_{q(\mathbf{x}(0)|\mathbf{x}(t))}[-\frac{1}{a(t)} \, \nabla_{\mathbf{x}(0)} \log q(\mathbf{x}(t)|\mathbf{x}(0))] \label{eq:gauss_sym} \\
                                             & \overset{\text{Bayes' rule}}{=} -\frac{1}{a(t)} \, \mathbb{E}_{q(\mathbf{x}(0)|\mathbf{x}(t))}\big[\nabla_{\mathbf{x}(0)} \log q(\mathbf{x}(0)|\mathbf{x}(t)) \nonumber \\
                                             &\quad + \nabla_{\mathbf{x}(0)} \log q_t(\mathbf{x}(t)) - \nabla_{\mathbf{x}(0)} \log p(\mathbf{x}(0))\big] \\
                                             & = -\frac{1}{a(t)} \left( \mathbb{E}_{q(\mathbf{x}(0)|\mathbf{x}(t))}[\nabla_{\mathbf{x}(0)} \log q(\mathbf{x}(0)|\mathbf{x}(t))] - \mathbb{E}_{q(\mathbf{x}(0)|\mathbf{x}(t))}[\nabla_{\mathbf{x}} \log p(\mathbf{x})] \right) \label{eq:zero_post} \\
                                             & = \frac{1}{a(t)} \mathbb{E}_{q(\mathbf{x}(0)|\mathbf{x}(t))}[\nabla_{\mathbf{x}} \log p(\mathbf{x})],
\end{align}

where in step~\eqref{eq:zero_post} we use:
\begin{align}
    \mathbb{E}_{q(\mathbf{x}(0) | \mathbf{x}(t))}[\nabla_{\mathbf{x}(0)} \log q(\mathbf{x}(0) | \mathbf{x}(t))] & = \int q(\mathbf{x}(0) | \mathbf{x}(t)) \frac{\nabla_{\mathbf{x}(0)} q(\mathbf{x}(0) | \mathbf{x}(t))}{q(\mathbf{x}(0) | \mathbf{x}(t))} \dd \mathbf{x}(0) \\
                                                                                                                & = \int \nabla_{\mathbf{x}(0)} q(\mathbf{x}(0) | \mathbf{x}(t)) \dd \mathbf{x}(0) \\
                                                                                                                & \overset{\text{Divergence theorem}}{=} 0,
    \label{eq:expec_post_zero}
\end{align}

and in step~\eqref{eq:gauss_sym} we use the Gaussian symmetry property:
\begin{align}
    q(\mathbf{x}(t)|\mathbf{x}(0))                             & = \mathcal{N} \left( \mathbf{x}(t); a(t) \mathbf{x}(0), b(t)^2 I \right), \\
    \nabla_{\mathbf{x}(0)} \log q(\mathbf{x}(t)|\mathbf{x}(0)) & = \frac{a(t)(\mathbf{x}(t) - a(t)\mathbf{x}(0))}{b(t)^2}, \\
    \nabla_{\mathbf{x}(t)} \log q(\mathbf{x}(t)|\mathbf{x}(0)) & = -\frac{\mathbf{x}(t) - a(t)\mathbf{x}(0)}{b(t)^2} \\
                                                               & = -\frac{1}{a(t)} \nabla_{\mathbf{x}(0)} \log q(\mathbf{x}(t)|\mathbf{x}(0)).
    \label{eq:grad_x0_proof}
\end{align}

\subsection{Derivation using integration by parts}
\label{app:proof_proposition1_B}
We start with the definition of the marginal $q_t(\mathbf{x}(t))$:
\begin{align*}
    \nabla_{\mathbf{x}(t)} \log q_t(\mathbf{x}(t))
     & = \frac{1}{q_t(\mathbf{x}(t))} \nabla_{\mathbf{x}(t)} \int q(\mathbf{x}(t) | \mathbf{x}(0)) p(\mathbf{x}(0)) \dd \mathbf{x}(0) \\
     & \overset{\text{(Leibniz rule)}}{=} \frac{1}{q_t(\mathbf{x}(t))} \int (\nabla_{\mathbf{x}(t)} q(\mathbf{x}(t) | \mathbf{x}(0))) p(\mathbf{x}(0)) \dd \mathbf{x}(0) \\
     & = \frac{1}{q_t(\mathbf{x}(t))} \int q(\mathbf{x}(t) | \mathbf{x}(0)) (\nabla_{\mathbf{x}(t)} \log q(\mathbf{x}(t) | \mathbf{x}(0))) p(\mathbf{x}(0)) \dd \mathbf{x}(0) \\
     & \left( = \mathbb{E}_{q(\mathbf{x}(0)|\mathbf{x}(t))}[\nabla_{\mathbf{x}(t)} \log q(\mathbf{x}(t)|\mathbf{x}(0))]  \quad \text{(DSI identity)}  \right) \\
     & \overset{\text{(Gaussian sym. Eq.~\eqref{eq:grad_x0_proof})}}{=} \frac{1}{q_t(\mathbf{x}(t))} \int q(\mathbf{x}(t) | \mathbf{x}(0)) \nonumber \\
     &\quad \times \left(-\frac{1}{a(t)} \nabla_{\mathbf{x}(0)} \log q(\mathbf{x}(t) | \mathbf{x}(0))\right) p(\mathbf{x}(0)) \dd \mathbf{x}(0) \\
     & = -\frac{1}{a(t)q_t(\mathbf{x}(t))} \int (\nabla_{\mathbf{x}(0)} q(\mathbf{x}(t) | \mathbf{x}(0))) p(\mathbf{x}(0)) \dd \mathbf{x}(0) \\
     & \overset{\text{(Integ. by parts)}}{=} \frac{1}{a(t)q_t(\mathbf{x}(t))} \int q(\mathbf{x}(t) | \mathbf{x}(0)) (\nabla_{\mathbf{x}(0)} p(\mathbf{x}(0))) \dd \mathbf{x}(0) \\
     & = \frac{1}{a(t)q_t(\mathbf{x}(t))} \int q(\mathbf{x}(t) | \mathbf{x}(0)) p(\mathbf{x}(0)) (\nabla_{\mathbf{x}(0)} \log p(\mathbf{x}(0))) \dd \mathbf{x}(0) \\
     & = \frac{1}{a(t)} \int q(\mathbf{x}(0)|\mathbf{x}(t)) \nabla_{\mathbf{x}(0)} \log p(\mathbf{x}(0)) \dd \mathbf{x}(0) \\
     & = \frac{1}{a(t)} \mathbb{E}_{q(\mathbf{x}(0)|\mathbf{x}(t))}[\nabla_{\mathbf{x}(0)} \log p(\mathbf{x}(0))].
\end{align*}

\section{Derivation of the interpolated estimator}
\label{app:derivation_interpolated}
We start with the control variate estimator from Eq.~\eqref{eq:cvsi_result}:
\begin{align}
    \nabla_{\mathbf{x}} \log q_t(\mathbf{x}) &= \mathbb{E}_{q(\mathbf{x}(0) | \mathbf{x}(t))} \bigg[ \frac{1}{a(t)} \, \nabla_{\mathbf{x}} \log p(\mathbf{x}) - c(\mathbf{x}(t), t) \nabla_{\mathbf{x}(0)} \log q(\mathbf{x}(0)|\mathbf{x}(t)) \bigg] \\
    &\overset{\text{Bayes' rule}}{=} \mathbb{E}_{q(\mathbf{x}(0) | \mathbf{x}(t))} \bigg[ \frac{1}{a(t)} \, \nabla_{\mathbf{x}} \log p(\mathbf{x}) \nonumber \\
    &\quad - c(\mathbf{x}(t), t) \big(\nabla_{\mathbf{x}} \log p(\mathbf{x}) + \nabla_{\mathbf{x}(0)} \log q(\mathbf{x}(t)|\mathbf{x}(0))\big) \bigg] \\
    &= \mathbb{E}_{q(\mathbf{x}(0) | \mathbf{x}(t))} \bigg[ \left(\frac{1}{a(t)} - c(\mathbf{x}(t), t)\right) \nabla_{\mathbf{x}} \log p(\mathbf{x}) \nonumber \\
    &\quad - c(\mathbf{x}(t), t) \nabla_{\mathbf{x}(0)} \log q(\mathbf{x}(t)|\mathbf{x}(0)) \bigg] \\
    &\overset{\text{Gauss. sym.}}{=} \mathbb{E}_{q(\mathbf{x}(0) | \mathbf{x}(t))} \bigg[ \left(\frac{1}{a(t)} - c(\mathbf{x}(t), t)\right) \nabla_{\mathbf{x}} \log p(\mathbf{x}) \nonumber \\
    &\quad + c(\mathbf{x}(t), t)a(t)\nabla_{\mathbf{x}(t)} \log q(\mathbf{x}(t)|\mathbf{x}(0)) \bigg] \\
    &= \frac{1 - c(\mathbf{x}(t), t) \, a(t)}{a(t)} \, \mathbb{E}_{q(\mathbf{x}(0) | \mathbf{x}(t))} \left[\nabla_{\mathbf{x}} \log p(\mathbf{x}) \right] \nonumber \\
    &\quad + c(\mathbf{x}(t), t)a(t) \mathbb{E}_{q(\mathbf{x}(0) | \mathbf{x}(t))} \left[\nabla_{\mathbf{x}(t)} \log q(\mathbf{x}(t)|\mathbf{x}(0)) \right].
\end{align}

\paragraph{Unifying previous approaches.}
\label{app:unif_prev}

The interpolated score formulation reveals that the standard DSI and TSI estimators, as well as previous mixing approaches \cite{de2024target}, are special cases of our framework (see Supplementary Table~\ref{tab:estimators_comparison}), corresponding to suboptimal choices for $\tilde{c}(\mathbf{x}(t), t)$, since $c^*(\mathbf{x}(t), t)$ is derived to strictly minimize the variance. We further highlight a critical distinction regarding the TSM Mode Mixing weighting proposed by De Bortoli \etal~\cite{de2024target}. It relies on the intra-mode variance $\sigma_{\text{mode}}^2$, a quantity that requires explicit knowledge of the ground truth mixture components. Since this information is generally unavailable in practice, it serves as a theoretical baseline. In contrast, our CVSI coefficient $c^*(\mathbf{x}(t), t)$ is based on tractable statistics, which does not require access to the ground truth.

Notably, the interpolation form with our optimal weight $\tilde{c}^*(\mathbf{x}(t), t) = c^*(\mathbf{x}(t), t)a(t)$ mathematically coincides with the mixture weights identified by Ko \etal~\cite{ko2025latent}.
However, while Ko \etal~\cite{ko2025latent} approach the problem by directly optimizing a linear interpolation between two estimators, our result emerges directly from the control variate framework, where the posterior score acts as the control variate.

\begin{table}[h]
    \centering
    \caption{Comparison of different score estimators as special cases of the Control Variate Score Identity (CVSI) based on the choice of the weighting function $\tilde{c}(\mathbf{x}(t), t)$. For the Target Score Matching (TSM) weightings \cite{de2024target}, $\sigma_{\text{data}}^2$ denotes the total data variance. Note that $\sigma_{\text{mode}}^2$ denotes the average intra-mode variance, which requires access to the ground truth mixture parameters.}
    \label{tab:estimators_comparison}
    \begin{tabular}{lcc}
        \toprule
        \textbf{Estimator} & \textbf{Weighting} $\tilde{c}(\mathbf{x}(t), t)$ & \textbf{Variance Behavior} \\
        \midrule
        Target Score Identity (TSI) & $0$ & Grows as $t \to 1$ \\
        Denoising Score Identity (DSI) & $1$ & Grows as $t \to 0$ \\
        TSM Global Mixing \cite{de2024target} & $\frac{b(t)^2}{b(t)^2 + a(t)^2 \sigma_{\text{data}}^2}$ & Optimal for Single Gaussian \\
        TSM Mode Mixing \cite{de2024target} & $\frac{b(t)^2}{b(t)^2 + a(t)^2 \sigma_{\text{mode}}^2}$ &  Heuristic for GMM (Requires GT) \\
        \textbf{CVSI (Ours)} & $c^*(\mathbf{x}(t), t)a(t)$ & \textbf{Minimal Variance} \\
        \bottomrule
    \end{tabular}
\end{table}

\section{Derivation of the optimal control coefficient}
\label{app:proof_optimal_c}

\subsection{Unbiased score estimator}
To show that the estimator in Eq.~\eqref{eq:cvsi_result} is unbiased, we need to show that
\begin{align*}
    \mathbb{E}\left[ g(\mathbf{x}(0)) - c(\mathbf{x}(t), t) \left(h(\mathbf{x}(0)) - \mathbb{E}[h(\mathbf{x}(0))]\right) \right]=\mathbb{E}_{q(\mathbf{x}(0) | \mathbf{x}(t))}[g(\mathbf{x}(0))].
\end{align*}
Starting with the proposed estimator, we have:
\begin{align*}
    &\mathbb{E}\big[ g(\mathbf{x}(0)) - c(\mathbf{x}(t), t) \left(h(\mathbf{x}(0)) - \mathbb{E}[h(\mathbf{x}(0))]\right) \big] \\
    &= \mathbb{E}[g(\mathbf{x}(0))] - \mathbb{E}\big[ c(\mathbf{x}(t), t) \left(h(\mathbf{x}(0)) - \mathbb{E}[h(\mathbf{x}(0))]\right) \big] \\
    &= \mathbb{E}[g(\mathbf{x}(0))] - c(\mathbf{x}(t), t) \cdot \big( \mathbb{E}[h(\mathbf{x}(0))] - \mathbb{E}[\mathbb{E}[h(\mathbf{x}(0))]] \big) \\
    &= \mathbb{E}[g(\mathbf{x}(0))] - c(\mathbf{x}(t), t) \cdot \big( \mathbb{E}[h(\mathbf{x}(0))] - \mathbb{E}[h(\mathbf{x}(0))] \big) \\
    &= \mathbb{E}[g(\mathbf{x}(0))],
\end{align*}
which proves that the estimator is unbiased.

\subsection{Derivation of \texorpdfstring{$c^*(\mathbf{x}(t), t)$}{c*(t)}}
\label{app:proof_c}

Let the control variate estimator for the integrand $g$ be $g_c = g - c(h - \mathbb{E}[h])$.
We want to find the value of $c$ that minimizes the variance of $g_c$.
The variance is given by:
\begin{align*}
    \Var(g_c) & = \Var(g - c(h - \mathbb{E}[h])) \\
              & \overset{\Var(A-B)}{=} \Var(g) + \Var(c(h - \mathbb{E}[h])) - 2\Cov(g, c(h - \mathbb{E}[h])) \\
              & = \Var(g) + c^2\Var(h - \mathbb{E}[h]) - 2c\Cov(g, h - \mathbb{E}[h]) \\
              & \overset{\mathbb{E}[h] \text{ is const.}}{=} \Var(g) + c^2\Var(h) - 2c\Cov(g, h).
\end{align*}
This expression for the variance is a quadratic function of $c$.
To find the minimum, we take the derivative with respect to $c$ and set it to zero:
\begin{align*}
    \frac{d}{dc} \Var(g_c)                      & = \frac{d}{dc} \left( \Var(g) + c^2\Var(h) - 2c\Cov(g, h) \right) \\
                                                & = 2c\Var(h) - 2\Cov(g, h). \\
    \text{Setting to zero} \implies 2c^*\Var(h) & = 2\Cov(g, h), \\
    c^* & = \frac{\Cov(g, h)}{\Var(h)}.
\end{align*}
Since the second derivative, $2\Var(h)$, is positive, this value of $c^*$ corresponds to a minimum, which completes the proof.

\subsection{Computationally tractable reformulation for \texorpdfstring{$c^*(\mathbf{x}(t), t)$}{c*(t)}}
\label{app:derivation_cstar}
We have:
\begin{itemize}
    \item $g = a(t)^{-1} \nabla_{\mathbf{x}} \log p(\mathbf{x})$
    \item $h = \nabla_{\mathbf{x}(0)} \log q(\mathbf{x}(0)|\mathbf{x}(t))$
    \item $k = \nabla_{\mathbf{x}(0)} \log q(\mathbf{x}(t)|\mathbf{x}(0))$
\end{itemize}
To make the derivation easy we consider the rescaled version $\tilde{g}(t)=a(t) \, g(t) = \nabla_{\mathbf{x}} \log p(\mathbf{x})$, and therefore we get:
\begin{align*}
    c^* & = \frac{1}{a(t)}\frac{\Cov(\tilde{g}, h)}{\Var(h)}.
\end{align*}
From the proof of Proposition 1, we know these terms are related by Bayes' rule such that $h = \tilde{g} + k$.
All expectations, variances, and covariances are taken with respect to the posterior, $q(\mathbf{x}(0)|\mathbf{x}(t))$.
Therefore, we can re-write $c^*(\mathbf{x}(t), t)$ as:

\paragraph{Numerator.}
\begin{align*}
    \text{Cov}(\tilde{g}, h) & = \text{Cov}(\tilde{g}, \tilde{g}+k) \\
                             & = \text{Cov}(\tilde{g}, \tilde{g}) + \text{Cov}(\tilde{g}, k) \\
                             & = \Var(\tilde{g}) + \text{Cov}(\tilde{g}, k).
\end{align*}

\paragraph{Denominator.}
\begin{align*}
    \Var(h) & = \Var(\tilde{g}+k) \\
            & = \Var(\tilde{g}) + \Var(k) + 2\text{Cov}(\tilde{g}, k).
\end{align*}

Therefore, we obtain the reformulated control coefficient:
\begin{equation}
\begin{split}
    c^*(\mathbf{x}(t), t) &= \frac{1}{a(t)} \big( \Var(\nabla_{\mathbf{x}} \log p(\mathbf{x})) + \Cov(\nabla_{\mathbf{x}} \log p(\mathbf{x}), \nabla_{\mathbf{x}(0)} \log q_{t|0}) \big) \\
    &\quad \times \big( \Var(\nabla_{\mathbf{x}} \log p(\mathbf{x})) + \Var(\nabla_{\mathbf{x}(0)} \log q_{t|0}) \\
    &\quad + 2\Cov(\nabla_{\mathbf{x}} \log p(\mathbf{x}), \nabla_{\mathbf{x}(0)} \log q_{t|0}) \big)^{-1},
\end{split}
\end{equation}

where $q_{t|0}=q(\mathbf{x}(t)|\mathbf{x}(0))$. Moreover, by using the identity:
\begin{align}
    \nabla_{\mathbf{x}(0)} \log q_{t|0} = -a(t) \nabla_{\mathbf{x}(t)} \log q_{t|0},
\end{align}

we get the final result:
\begin{align}
    c^*(\mathbf{x}(t), t) &= \frac{1}{a(t)} \big( \Var(\nabla_{\mathbf{x}} \log p(\mathbf{x})) - a(t) \Cov(\nabla_{\mathbf{x}} \log p(\mathbf{x}), \nabla_{\mathbf{x}(t)} \log q_{t|0}) \big) \nonumber \\
    &\quad \times \big( \Var(\nabla_{\mathbf{x}} \log p(\mathbf{x})) + a(t)^2 \Var(\nabla_{\mathbf{x}(t)} \log q_{t|0}) \nonumber \\
    &\quad - 2 a(t) \Cov(\nabla_{\mathbf{x}} \log p(\mathbf{x}), \nabla_{\mathbf{x}(t)} \log q_{t|0}) \big)^{-1} \nonumber \\
    &= \big( \Var(\nabla_{\mathbf{x}} \log p(\mathbf{x})) - a(t) \Cov(\nabla_{\mathbf{x}} \log p(\mathbf{x}), \nabla_{\mathbf{x}(t)} \log q_{t|0}) \big) \nonumber \\
    &\quad \times \big( a(t) \Var(\nabla_{\mathbf{x}} \log p(\mathbf{x})) + a(t)^3 \Var(\nabla_{\mathbf{x}(t)} \log q_{t|0}) \nonumber \\
    &\quad - 2 a(t)^2 \Cov(\nabla_{\mathbf{x}} \log p(\mathbf{x}), \nabla_{\mathbf{x}(t)} \log q_{t|0}) \big)^{-1}.
\end{align}

\paragraph{Boltzmann distributions.}

In the case of a Boltzmann distribution defined by an energy function $E(\mathbf{x})$ and temperature parameter $\tau$, such that $p(\mathbf{x}) \propto \exp(-E(\mathbf{x})/\tau)$, the target score is given by $\nabla_{\mathbf{x}} \log p(\mathbf{x}) = -\nabla_{\mathbf{x}} E(\mathbf{x}) / \tau$.
Substituting the analytical score of the perturbation kernel $\nabla_{\mathbf{x}(t)} \log q(\mathbf{x}(t)|\mathbf{x}(0)) = \frac{a(t) \, \mathbf{x}(0) - \mathbf{x}(t) }{b(t)^2}$, the optimal coefficient becomes:
\begin{equation}
\begin{split}
    c^*(\mathbf{x}(t), t) &= \left( \frac{1}{\tau^2} \Var(\nabla E) + \frac{a(t)^2}{\tau b(t)^2} \Cov(\nabla E, \mathbf{x}(0)) \right) \\
    &\quad \times \left( \frac{a(t)}{\tau^2} \Var(\nabla E) + \frac{a(t)^5}{b(t)^4} \Var(\mathbf{x}(0)) + \frac{2 a(t)^3}{\tau b(t)^2} \Cov(\nabla E, \mathbf{x}(0)) \right)^{-1},
\end{split}
\end{equation}
where $\nabla E$ denotes $\nabla_{\mathbf{x}(0)} E(\mathbf{x}(0))$ and all statistics are taken w.r.t. $q(\mathbf{x}(0)|\mathbf{x}(t))$. This explicit form allows for the direct computation of optimal variance reduction in physical systems where the energy function is known.

\section{Derivation and reframing of previous methods}
\subsection{iDEM}
\label{app:idem}
Let the scaled variable be $\tilde{\mathbf{x}}(t) = \mathbf{x}(t)/a(t)$, and the SNR ratio be $\gamma(t) = a(t)/b(t)$.
Using the proposal distribution $\pi_{\text{iDEM}}(\mathbf{x}(0)|\tilde{\mathbf{x}}(t))=\mathcal{N}(\mathbf{x}(0); \tilde{\mathbf{x}}(t), \gamma(t)^{-2})$, the iDEM estimator \cite{akhound2024iterated} is defined as:
\begin{align}
    \nabla_{\tilde{\mathbf{x}}(t)} \log \tilde{q}_t(\tilde{\mathbf{x}}(t))
     & = \nabla_{\tilde{\mathbf{x}}(t)} \log \mathbb{E}_{\pi_{\text{iDEM}}(\mathbf{x}(0)|\tilde{\mathbf{x}}(t))} \left[p(\mathbf{x}(0))\right] \\
     & = \frac{\mathbb{E}_{\pi_{\text{iDEM}}(\mathbf{x}(0)|\tilde{\mathbf{x}}(t))}\left[\nabla_{\mathbf{x}(0)}p(\mathbf{x}(0))\right]}{\mathbb{E}_{\pi_{\text{iDEM}}(\mathbf{x}(0)|\tilde{\mathbf{x}}(t))}\left[p(\mathbf{x}(0))\right]},
\end{align}
which can be developed as follows:
\begin{align}
     & \frac{\mathbb{E}_{\pi_{\text{iDEM}}(\mathbf{x}(0)|\tilde{\mathbf{x}}(t))}\left[\nabla_{\mathbf{x}(0)}p(\mathbf{x}(0))\right]}{\mathbb{E}_{\pi_{\text{iDEM}}(\mathbf{x}(0)|\tilde{\mathbf{x}}(t))}\left[p(\mathbf{x}(0))\right]} \notag \\
     & =  \frac{1}{\tilde{q}_t(\tilde{\mathbf{x}}(t))} \int \nabla_{\mathbf{x}(0)}p(\mathbf{x}(0)) \pi_{\text{iDEM}}(\mathbf{x}(0)|\tilde{\mathbf{x}}(t)) \dd \mathbf{x}(0) \\
     & = \int \frac{p(\mathbf{x}(0))}{\tilde{q}_t(\tilde{\mathbf{x}}(t))} \nabla_{\mathbf{x}(0)} \log p(\mathbf{x}(0)) \pi_{\text{iDEM}}(\mathbf{x}(0)|\tilde{\mathbf{x}}(t)) \dd \mathbf{x}(0) \\
     & =  \mathbb{E}_{\pi_{\text{iDEM}}(\mathbf{x}(0)|\tilde{\mathbf{x}}(t))} \left[\frac{p(\mathbf{x}(0))}{\tilde{q}_t(\tilde{\mathbf{x}}(t))} \nabla_{\mathbf{x}(0)} \log p(\mathbf{x}(0)) \right] \\
     &= \mathbb{E}_{\pi_{\text{iDEM}}} \bigg[ \frac{p(\mathbf{x}(0)) \pi_{\text{iDEM}}(\mathbf{x}(0)|\tilde{\mathbf{x}}(t))}{\tilde{q}_t(\tilde{\mathbf{x}}(t)) \pi_{\text{iDEM}}(\mathbf{x}(0)|\tilde{\mathbf{x}}(t))} \nabla_{\mathbf{x}(0)} \log p(\mathbf{x}(0)) \bigg] \nonumber \\
     &= \mathbb{E}_{\pi_{\text{iDEM}}} \bigg[ \frac{p(\mathbf{x}(0)) \tilde{q}(\tilde{\mathbf{x}}(t)|\mathbf{x}(0))}{\tilde{q}_t(\tilde{\mathbf{x}}(t)) \pi_{\text{iDEM}}(\mathbf{x}(0)|\tilde{\mathbf{x}}(t))} \nabla_{\mathbf{x}(0)} \log p(\mathbf{x}(0)) \bigg] \\
     & = \mathbb{E}_{\pi_{\text{iDEM}}(\mathbf{x}(0)|\tilde{\mathbf{x}}(t))} \left[ \frac{\tilde{q}(\mathbf{x}(0)|\tilde{\mathbf{x}}(t))}{\pi_{\text{iDEM}}(\mathbf{x}(0)|\tilde{\mathbf{x}}(t))} \nabla_{\mathbf{x}(0)} \log p(\mathbf{x}(0)) \right] \\
     & \overset{\tilde{q}(\mathbf{x}(0)|\tilde{\mathbf{x}}(t)) = q(\mathbf{x}(0)|\mathbf{x}(t))}{=} \mathbb{E}_{\pi_{\text{iDEM}}(\mathbf{x}(0)|\tilde{\mathbf{x}}(t))} \left[ \frac{q(\mathbf{x}(0)|\mathbf{x}(t))}{\pi_{\text{iDEM}}(\mathbf{x}(0)|\tilde{\mathbf{x}}(t))} \nabla_{\mathbf{x}(0)}\log p(\mathbf{x}(0)) \right] \label{eq:is_idem} \\
     & =  \mathbb{E}_{q(\mathbf{x}(0)|\mathbf{x}(t))} \left[\nabla_{\mathbf{x}(0)} \log p(\mathbf{x}(0)) \right],
\end{align}
where we use the following equalities:
\begin{align}
    \pi_{\text{iDEM}}(\mathbf{x}(0)|\tilde{\mathbf{x}}(t))=\mathcal{N}(\mathbf{x}(0); \tilde{\mathbf{x}}(t), \gamma(t)^{-2})= \tilde{q}(\tilde{\mathbf{x}}(t)|\mathbf{x}(0)),
\end{align}
and
\begin{align}
    \tilde{q}(\mathbf{x}(0)|\tilde{\mathbf{x}}(t)) & = \frac{\tilde{q}(\tilde{\mathbf{x}}(t)|\mathbf{x}(0))p(\mathbf{x}(0))}{\tilde{q}_t(\tilde{\mathbf{x}}(t))} \\
                                                   & \overset{\text{(Change of var.)}}{=} \frac{a(t)^D q(\mathbf{x}(t)|\mathbf{x}(0))p(\mathbf{x}(0)) }{a(t)^D q_t(\mathbf{x}(t))} \\
                                                   & = \frac{q(\mathbf{x}(t)|\mathbf{x}(0))p(\mathbf{x}(0)) }{q_t(\mathbf{x}(t))} \\
                                                   & = q(\mathbf{x}(0)|\mathbf{x}(t)),
\end{align}
where $D$ is the dimensionality of the data.
While the above derivation holds for any schedule, the iDEM estimator is primarily associated with the Variance Exploding (VE) framework.
However, we can relate the estimator for the scaled variable $\tilde{\mathbf{x}}(t)$ to the unscaled original variable $\mathbf{x}(t)$ as follows:
\begin{align}
    \nabla_{\mathbf{x}(t)} \log q_t(\mathbf{x}(t))
     & =  \nabla_{\mathbf{x}(t)} \log \left( \frac{1}{a(t)^D} \tilde{q}_t(\tilde{\mathbf{x}}(t)) \right) \\
     & =  \nabla_{\mathbf{x}(t)} \log \tilde{q}_t(\tilde{\mathbf{x}}(t)) \\
     & =  \frac{1}{a(t)} \nabla_{\tilde{\mathbf{x}}(t)} \log \tilde{q}_t(\tilde{\mathbf{x}}(t)) \\
     &= \frac{1}{a(t)} \mathbb{E}_{\pi_{\text{iDEM}}} \bigg[ \frac{q(\mathbf{x}(0)|\mathbf{x}(t))}{\pi_{\text{iDEM}}(\mathbf{x}(0)|\tilde{\mathbf{x}}(t))} \nabla_{\mathbf{x}(0)}\log p_0(\mathbf{x}(0)) \bigg] \label{eq:is_idem_general} \\
     & =  \frac{1}{a(t)}  \mathbb{E}_{q(\mathbf{x}(0)|\mathbf{x}(t))} \left[\nabla_{\mathbf{x}(0)} \log p_0(\mathbf{x}(0)) \right].
\end{align}

Eq.~\eqref{eq:is_idem} and its unscaled version Eq.~\eqref{eq:is_idem_general} show that the estimator used in iDEM is an importance sampling (IS) estimator for the target score identity from Eq.~\eqref{eq:methods_tsi}, with importance weights:
\begin{align}
    w_{\text{iDEM}}(\mathbf{x}(t), t) & = \frac{p(\mathbf{x}(0))}{\tilde{q}_t(\tilde{\mathbf{x}}(t))} \\
                                      & = \frac{q(\mathbf{x}(0)|\mathbf{x}(t))}{\pi_{\text{iDEM}}(\mathbf{x}(0)|\tilde{\mathbf{x}}(t))} \\
                                      & = \frac{q(\mathbf{x}(0)|\mathbf{x}(t))}{\tilde{q}(\tilde{\mathbf{x}}(t)|\mathbf{x}(0))} \\
                                      & \overset{\text{(Change of var.)}}{=} \frac{q(\mathbf{x}(0)|\mathbf{x}(t))}{a(t)^D \, q(\mathbf{x}(t)|\mathbf{x}(0))} \\
                                      & = \frac{p(\mathbf{x}(0))}{a(t)^D \, q_t(\mathbf{x}(t))}.
\end{align}
The variance of this importance weight is lowest when the proposal distribution $\pi_{\text{iDEM}}(\mathbf{x}(0)|\tilde{\mathbf{x}}(t))$ is a good match for the target posterior $q(\mathbf{x}(0)|\mathbf{x}(t))$, which occurs as $t \to 0$, where the distributions collapse into sharp Gaussians centered around $\mathbf{x}(t)$.
Conversely, the variance is very high when $t$ is large.

\section{Analytical solution for diffused multivariate Gaussian Mixture Models}
\label{app:analytical_gmm}

We consider a more general case of De Bortoli \etal~\cite{de2024target}, where the initial data distribution is modeled as a multivariate Gaussian Mixture Model (GMM) in $\mathbb{R}^d$:
\begin{equation}
    p(\mathbf{x}(0)) = \sum_{i=1}^{N} \pi^i \mathcal{N}(\mathbf{x}(0); \boldsymbol{\mu}^i, \boldsymbol{\Sigma}^i).
\end{equation}
We use a superscript $i$ (e.g., $\boldsymbol{\mu}^i$) to denote the $i$-th component of the mixture model, in order to reserve subscripts for vector or matrix element indexing.

\subsection{The marginal Distribution $q(\mathbf{x}(t))$}
To find the marginal distribution $q(\mathbf{x}(t))$, we integrate over all possible initial states $\mathbf{x}(0)$:
\begin{align*}
    q(\mathbf{x}(t)) & = \int q(\mathbf{x}(t)|\mathbf{x}(0)) p(\mathbf{x}(0)) d\mathbf{x}(0) \\
                     & = \int \mathcal{N}(\mathbf{x}(t); a(t)\mathbf{x}(0), b(t)^2\mathbf{I}) \left( \sum_{i=1}^{N} \pi^i \mathcal{N}(\mathbf{x}(0); \boldsymbol{\mu}^i, \boldsymbol{\Sigma}^i) \right) \dd \mathbf{x}(0) \\
                     & = \sum_{i=1}^{N} \pi^i \int \mathcal{N}(\mathbf{x}(t); a(t)\mathbf{x}(0), b(t)^2\mathbf{I}) \, \mathcal{N}(\mathbf{x}(0); \boldsymbol{\mu}^i, \boldsymbol{\Sigma}^i) \dd \mathbf{x}(0).
\end{align*}
The integral represents the convolution of two multivariate Gaussians, which results in another multivariate Gaussian.
The new mean vector and covariance for each component $i$ are:
\begin{align*}
    \mathbb{E}[\mathbf{x}(t)] & = a(t)\mathbb{E}[\mathbf{x}(0)] + \mathbb{E}[\mathbf{w}] = a(t)\boldsymbol{\mu}^i, \\
    \Cov(\mathbf{x}(t))        & = a(t)^2 \Cov(\mathbf{x}(0)) + \Cov(\mathbf{w}) = a(t)^2\boldsymbol{\Sigma}^i + b(t)^2\mathbf{I},
\end{align*}
where $\mathbf{w} \sim \mathcal{N}(\mathbf{0}, b(t)^2 \mathbf{I})$, and assuming independence between $\mathbf{x}(0)$ and $\mathbf{w}$. Therefore, we get:
\begin{align}
    q_t(\mathbf{x})                             & = \sum_{i=1}^{N} \pi^i \mathcal{N}(\mathbf{x}; \boldsymbol{\mu}^i(t), \boldsymbol{\Sigma}^i(t)), \\
    \text{where} \quad \boldsymbol{\mu}^i(t)  & = a(t)\boldsymbol{\mu}^i, \\
    \text{and} \quad \boldsymbol{\Sigma}^i(t) & = a(t)^2\boldsymbol{\Sigma}^i + b(t)^2\mathbf{I}.
\end{align}

\subsection{The posterior distribution $q(\mathbf{x}(0)|\mathbf{x}(t))$}
\label{app:analytical_gmm_reverse}
We use Bayes' theorem:
\begin{align*}
    q(\mathbf{x}(0)|\mathbf{x}(t)) & = \frac{q(\mathbf{x}(t)|\mathbf{x}(0)) p(\mathbf{x}(0))}{q(\mathbf{x}(t))}.
\end{align*}
The posterior is proportional to the product of the likelihood and the prior, which is the product of two Gaussians for a single component $i$:
\begin{equation*}
    \mathcal{N}(\mathbf{x}(t); a(t)\mathbf{x}(0), b(t)^2 \mathbf{I}) \times \mathcal{N}(\mathbf{x}(0); \boldsymbol{\mu}^i, \boldsymbol{\Sigma}^i).
\end{equation*}
Therefore, by applying the standard update rules from the Kalman filtering framework for linear-Gaussian systems, we get the posterior covariance for each component $i$:
\begin{align}
    \boldsymbol{\Gamma}^i(t) & = \left( (\boldsymbol{\Sigma}^i)^{-1} + \frac{a(t)^2}{b(t)^2}\mathbf{I} \right)^{-1},
\end{align}
and mean:
\begin{align}
    \boldsymbol{\nu}^i(t) & = \boldsymbol{\Gamma}^i(t) \left( \frac{a(t)}{b(t)^2}\mathbf{x}(t) + (\boldsymbol{\Sigma}^i)^{-1}\boldsymbol{\mu}^i \right).
\end{align}
Given the above posterior means and covariances, we get the posterior GMM:
\begin{equation}
    q(\mathbf{x}(0)|\mathbf{x}(t)) = \sum_{i=1}^{N} \pi^i(t) \mathcal{N}(\mathbf{x}(0); \boldsymbol{\nu}^i(t), \boldsymbol{\Gamma}^i(t)),
\end{equation}
with the time-dependent mixture weights $\pi^i(t)$ representing the posterior probabilities of belonging to component $i$ at time $t$:
\begin{equation}
    \pi^i(t) = \frac{\pi^i \mathcal{N}(\mathbf{x}(t); \boldsymbol{\mu}^i(t), \boldsymbol{\Sigma}^i(t))}{\sum_{j=1}^{N} \pi^j \mathcal{N}(\mathbf{x}(t); \boldsymbol{\mu}^j(t), \boldsymbol{\Sigma}^j(t))}.
\end{equation}

\section{Analysis of time-dependent score estimation error}
\label{app:gaussian_error_analysis}

Let the approximated score be $\mathbf{s}_\theta(\mathbf{x}(t), t) = \nabla_{\mathbf{x}(t)} \log q_t(\mathbf{x}(t)) + \boldsymbol{\xi}(t)$, where $\boldsymbol{\xi}(t)$ represents approximation noise, e.g., during the first epochs of iDEM training from the replay buffer. We model this error as a time-dependent Gaussian process $\boldsymbol{\xi}(t) \sim \mathcal{N}\left(\mathbf{0}, \sigma_{\text{err}}^2(t) \mathbf{I}\right)$, which can be shown to hold empirically.

Substituting into the reverse SDE, we obtain:
\begin{align}
    \dd \tilde{\mathbf{x}}(t) &= \left[ f(t) \tilde{\mathbf{x}}(t) - \frac{1+\lambda^2}{2} {g(t)}^2 \nabla_{\tilde{\mathbf{x}}(t)} \log q_t(\tilde{\mathbf{x}}(t))\right] \dd \tilde{t} + \lambda {g(t)} \, \dd \tilde{\mathbf{w}}(t) \nonumber \\
    &= \left[ f(t) \tilde{\mathbf{x}}(t) - \frac{1+\lambda^2}{2} {g(t)}^2 \left( \mathbf{s}_\theta(\tilde{\mathbf{x}}(t), t) - \boldsymbol{\xi}(t) \right) \right] \dd \tilde{t} + \lambda {g(t)} \, \dd \tilde{\mathbf{w}}(t) \nonumber \\
    &= \underbrace{\left[ f(t) \tilde{\mathbf{x}}(t) - \frac{1+\lambda^2}{2} {g(t)}^2 \mathbf{s}_\theta(\tilde{\mathbf{x}}(t), t) \right] \dd \tilde{t}}_{\text{Modeled Reverse Drift}} \nonumber \\
    &\quad + \underbrace{\frac{1+\lambda^2}{2} {g(t)}^2 \boldsymbol{\xi}(t) \dd \tilde{t}}_{\text{Error Drift Term}} + \underbrace{\lambda {g(t)} \, \dd \tilde{\mathbf{w}}(t)}_{\text{Diffusion}}.
    \label{eq:perturbed_sde}
\end{align}

\paragraph{Discrete time case (finite step $\Delta t$).}
In numerical solvers like Euler-Maruyama, the process is discretized with step size $\Delta t$. Both the error term and the Wiener process act as Gaussian perturbations to the trajectory update. Therefore, the variance of the update for a single step at time $t$, conditioned on $\tilde{\mathbf{x}}(t)$, is:
\begin{align}
    \Var(\Delta \tilde{\mathbf{x}}_t | \tilde{\mathbf{x}}(t)) &\approx \Var\left( \lambda g(t) \Delta \mathbf{w} \right) + \Var\left( \frac{1+\lambda^2}{2} g(t)^2 \boldsymbol{\xi}(t) \Delta t \right) \nonumber \\
    &= \lambda^2 g(t)^2 \Delta t \nonumber \\
    &\quad + \left(\frac{1+\lambda^2}{2}\right)^2 g(t)^4 \sigma_{\text{err}}^2(t) (\Delta t)^2.
\end{align}
Even if the score error $\sigma_{\text{err}}(t)$ is small at high noise levels ($t \to 1$), the large magnitude of $g(t)$ may amplify this term.

\paragraph{Continuous time case (limit $\Delta t \to 0$).}
In the continuous limit, we cannot simply merge the error term into the Wiener process variance. The diffusion term scales with $\sqrt{\Delta t}$, while the error drift scales with $\Delta t$. Examining the ratio of their variances as $\Delta t \to 0$:
\begin{align}
 \lim_{\Delta t \to 0} \frac{ \left[ \frac{1+\lambda^2}{2} g(t)^2 \sigma_{\text{err}}(t) \right]^2 (\Delta t)^2 }{ \lambda^2 g(t)^2 \Delta t } 
    &= \lim_{\Delta t \to 0} \left[ \frac{(1+\lambda^2)^2 g(t)^2 \sigma_{\text{err}}^2(t)}{4 \lambda^2} \right] \cdot \Delta t = 0.
\end{align}
Consequently, in the theoretical continuous limit, the Gaussian training error acts as a perturbation of the drift rather than a modification of the diffusion coefficient, and its contribution to the variance vanishes relative to the infinite variation of the diffusion term.

\paragraph{Effective variance and diffusion compensation.}
While theoretically distinct in the continuous limit, for a fixed simulation step size $\Delta t$, the error acts as an effective increase in the temperature of the sampling process. The total inflated noise scale $\lambda_{\text{inflated}}$ becomes:
\begin{equation}
    \lambda_{\text{inflated}}(t, \Delta t) = \sqrt{\lambda^2 + \left( \frac{1+\lambda^2}{2} \right)^2 g(t)^2 \sigma_{\text{err}}^2(t) \Delta t}.
    \label{eq:lambda_inflated}
\end{equation}
To mitigate this temperature inflation and maintain the target variance dictated by $\lambda^2$, one must scale down the explicit noise injected by the Wiener process. By balancing the variances, we derive the compensated effective diffusion scale $\lambda_\mathrm{eff}$:
\begin{equation}
    \lambda_{\text{eff}}(t, \Delta t) = \sqrt{\lambda^2 - \left( \frac{1+\lambda^2}{2} \right)^2 g(t)^2 \sigma_{\text{err}}^2(t) \Delta t}.
    \label{eq:supp_lambda_eff}
\end{equation}
Consequently, the actual reverse SDE implemented in practice uses this reduced coefficient alongside standard Wiener noise:
\begin{equation}
\begin{split}
    \dd \tilde{\mathbf{x}}(t) &= \left[ f(t) \tilde{\mathbf{x}}(t) - \frac{1+\lambda^2}{2} {g(t)}^2 \mathbf{s}_\theta(\tilde{\mathbf{x}}(t), t) \right] \dd \tilde{t} \\
    &\quad + \lambda_{\text{eff}}(t, \Delta t) \, g(t) \, \dd \tilde{\mathbf{w}}(t).
\end{split}
\label{eq:practical_sampler_sde}
\end{equation}
Crucially, adjusting $\lambda_\mathrm{eff}$ only corrects the magnitude (variance inflation) of the error; it does not correct the direction (deviation from the data manifold) caused by the error pushing the sample off the true probability flow. Furthermore, this variance correction is strictly limited: if the error variance $\sigma_{\text{err}}^2(t)$ is too large, the term under the square root in Eq.~\eqref{eq:supp_lambda_eff} becomes negative. This physically indicates that the approximation noise alone injects more variance than the target $\lambda^2$, rendering exact compensation impossible.
This strict bound explains why estimators with high variance, like TSI, might not benefit from this corrective approach.

\paragraph{The data-free sampler learning case.}
In data-free iterative frameworks like iDEM, this effective temperature inflation causes noisy trajectories that populate the replay buffer, creating a destructive feedback loop that exacerbates the score approximation error $\boldsymbol{\xi}(t)$ in subsequent epochs. 

To mitigate this, we must account for the stochasticity introduced by the error by replacing the standard $\lambda$ coefficient on the Wiener process with the reduced $\lambda_\mathrm{eff}$ (as shown in Eq.~\eqref{eq:practical_sampler_sde}), \textit{without} scaling the modeled reverse drift. This intentionally breaks the symmetry defined by $\lambda$ in the standard reverse SDE (Eq.~\eqref{eq:methods_rev_sde}). 

Although it is possible to use the exact analytical expression in Eq.~\eqref{eq:supp_lambda_eff} by estimating the variance of the score error $\sigma^2_{\text{err}}(t)$ at each training and time step, we empirically found that manually setting $\lambda_\mathrm{eff}$ to a constant low value, e.g., $\lambda_\mathrm{eff}=0.5$, is often sufficient to break the feedback loop, stabilize training, and improve sampling.

Note that scaling down the diffusion term independently of the drift alters the equilibrium distribution of the ideal SDE. If the score estimation were perfect ($\boldsymbol{\xi}(t) = 0$), using a low $\lambda_\mathrm{eff} < \lambda$ would result in low-temperature sampling from an exponentiated target distribution $\mathbf{x}(0) \sim p^w(\mathbf{x}(0))$ where $w > 1$. However, in our practical setting, this reduced $\lambda_\mathrm{eff}$ safely counteracts the temperature inflation related to the error, allowing the reverse process to more faithfully recover the true distribution $p(\mathbf{x}(0))$.

\section{Derivation of the exact posterior sampling in the Gaussian--Bernoulli RBM}
\label{app:gbrbm_derivations}

\subsection{The Gaussian--Bernoulli RBM bipartite structure}
The Gaussian--Bernoulli RBM~\cite{welling2004exponential} targeted in our image experiments models the joint probability of continuous visible units $\mathbf{v}\in\mathbb{R}^D$ and binary latent units $\mathbf{h}\in\{0,1\}^M$ via the energy function:
\begin{equation}
    E(\mathbf{v},\mathbf{h}) = \frac{\lVert \mathbf{v}-\mathbf{d}_v\rVert^2}{2\sigma^2} - \frac{\mathbf{h}^\top W \mathbf{v}}{\sigma^2} - \mathbf{d}_h^\top \mathbf{h},
    \label{eq:gbrbm_energy_app}
\end{equation}
with weight matrix $W\in\mathbb{R}^{M\times D}$, fixed visible noise scale $\sigma>0$, and $d_{h,j}$ representing the $j$-th entry of the offset $\mathbf{d}_h$. The bilinear coupling in Eq.~\eqref{eq:gbrbm_energy_app} gives the joint distribution $p(\mathbf{v},\mathbf{h})\propto\exp(-E(\mathbf{v},\mathbf{h}))$ a standard bipartite structure:
\begin{align}
    p(\mathbf{h}|\mathbf{v}) &= \prod_{j=1}^{M} \mathrm{Bernoulli}\!\left(h_j;\, \sigm\!\left(d_{h,j} + \frac{(W\mathbf{v})_j}{\sigma^2}\right)\right), \label{eq:h_given_v_app} \\
    p(\mathbf{v}|\mathbf{h}) &= \mathcal{N}\big(\mathbf{v};\, \mathbf{d}_v + W^\top\mathbf{h},\, \sigma^2\mathbf{I}\big). \label{eq:v_given_h_app}
\end{align}
The target marginal distribution is $p(\mathbf{v}) = \sum_{\mathbf{h}\in\{0,1\}^M} p(\mathbf{v},\mathbf{h}) \propto \exp(-F(\mathbf{v}))$. By summing out the binary latent units, which are conditionally independent given $\mathbf{v}$, the unnormalized free energy $F(\mathbf{v})$ is obtained in closed form:
\begin{equation}
    F(\mathbf{v}) = \frac{\lVert \mathbf{v}-\mathbf{d}_v\rVert^2}{2\sigma^2} - \sum_{j=1}^{M} \mathrm{softplus}\!\left(d_{h,j} + \frac{(W\mathbf{v})_j}{\sigma^2}\right).
    \label{eq:free_energy_app}
\end{equation}
While the normalizing constant $Z=\sum_{\mathbf{h}}\int \exp(-E(\mathbf{v},\mathbf{h}))\,\dd\mathbf{v}$ remains intractable, preventing direct sampling from $p(\mathbf{v})$, the closed-form nature of $F(\mathbf{v})$ ensures the TSI integrand $\nabla_{\mathbf{v}}\log p(\mathbf{v}) = -\nabla_{\mathbf{v}} F(\mathbf{v})$ is analytically available for any drawn sample.

\subsection{The augmented diffusion posterior}
\label{app:proof_gbrbm_conjugacy}

For a fixed diffused state $\mathbf{x}(t)$, the diffusion posterior over the visible units is $q(\mathbf{x}(0)|\mathbf{x}(t)) \equiv q(\mathbf{v}|\mathbf{x}(t)) \propto p(\mathbf{v})\, \mathcal{N}(\mathbf{x}(t); a(t)\mathbf{v}, b(t)^2\mathbf{I})$. To circumvent its intractability, we reintroduce $\mathbf{h}$ to form the augmented diffusion posterior:
\begin{equation}
    q(\mathbf{v},\mathbf{h}|\mathbf{x}(t)) \;\propto\; \exp\big(-E(\mathbf{v},\mathbf{h})\big) \, \mathcal{N}\big(\mathbf{x}(t);\, a(t)\mathbf{v},\, b(t)^2\mathbf{I}\big).
    \label{eq:tethered_joint_app}
\end{equation}
For every $t\in(0,1]$ and every diffusion state $\mathbf{x}(t)\in\mathbb{R}^D$, the augmented diffusion posterior $q(\mathbf{v},\mathbf{h}|\mathbf{x}(t))$ maintains closed-form full conditionals for exact block-Gibbs sampling. Writing out the negative logarithm of the augmented posterior in Eq.~\eqref{eq:tethered_joint_app} gives:
\begin{equation}
    -\log q(\mathbf{v},\mathbf{h}|\mathbf{x}(t)) = \frac{\lVert \mathbf{v}-\mathbf{d}_v\rVert^2}{2\sigma^2} - \frac{\mathbf{h}^\top W \mathbf{v}}{\sigma^2} - \mathbf{d}_h^\top\mathbf{h} + \frac{\lVert \mathbf{x}(t)-a(t)\mathbf{v}\rVert^2}{2b(t)^2} + \mathrm{const}.
    \label{eq:tethered_exponent_app}
\end{equation}

\paragraph{Conditional over $\mathbf{h}$.} 
By isolating the terms dependent on $\mathbf{h}$ in Eq.~\eqref{eq:tethered_exponent_app}, we obtain the unnormalized conditional density for $\mathbf{h}$:
\begin{align}
    -\log q(\mathbf{h}|\mathbf{v},\mathbf{x}(t)) &= -\mathbf{h}^\top \left( \mathbf{d}_h + \frac{W\mathbf{v}}{\sigma^2} \right) + C(\mathbf{v}, \mathbf{x}(t)),
\end{align}
where $C(\mathbf{v}, \mathbf{x}(t))$ acts as the normalizing constant. Because the diffusion likelihood term $\lVert \mathbf{x}(t)-a(t)\mathbf{v}\rVert^2/(2b(t)^2)$ strictly depends on $\mathbf{v}$, it is completely absorbed into $C$. Exponentiating this expression recovers the exact factorial Bernoulli distribution of the unperturbed energy model:
\begin{equation}
    q(\mathbf{h}|\mathbf{v},\mathbf{x}(t)) \propto \exp\left( \mathbf{h}^\top \left( \mathbf{d}_h + \frac{W\mathbf{v}}{\sigma^2} \right) \right) \propto p(\mathbf{h}|\mathbf{v}).
\end{equation}
Thus, the conditional over $\mathbf{h}$ remains completely unaffected by the diffusion process, resulting in $q(\mathbf{h}|\mathbf{v},\mathbf{x}(t)) = p(\mathbf{h}|\mathbf{v})$, which matches Eq.~\eqref{eq:h_given_v_tethered}.

\paragraph{Conditional over $\mathbf{v}$.} 
Similarly, isolating the terms dependent on $\mathbf{v}$ in Eq.~\eqref{eq:tethered_exponent_app} for a fixed $\mathbf{h}$ gives the unnormalized conditional density:
\begin{align}
    -\log q(\mathbf{v}|\mathbf{h},\mathbf{x}(t)) &= \frac{\lVert \mathbf{v}-\mathbf{d}_v\rVert^2}{2\sigma^2} - \frac{\mathbf{h}^\top W\mathbf{v}}{\sigma^2} + \frac{\lVert \mathbf{x}(t)-a(t)\mathbf{v}\rVert^2}{2b(t)^2} + C'(\mathbf{h}, \mathbf{x}(t)) \nonumber \\
    &= \frac{\lVert \mathbf{v}-(\mathbf{d}_v+W^\top\mathbf{h})\rVert^2}{2\sigma^2} + \frac{\lVert \mathbf{x}(t)-a(t)\mathbf{v}\rVert^2}{2b(t)^2} + C''(\mathbf{h}, \mathbf{x}(t)), \label{eq:v_conditional_expanded}
\end{align}
where completing the square for the cross term $-\mathbf{h}^\top W\mathbf{v}/\sigma^2$ introduces constants dependent on $\mathbf{h}$ that are absorbed into $C''$. To identify the resulting distribution, we expand Eq.~\eqref{eq:v_conditional_expanded} and collect the second- and first-order terms with respect to $\mathbf{v}$:
\begin{align}
    -\log q(\mathbf{v}|\mathbf{h},\mathbf{x}(t)) &= \frac{1}{2} \mathbf{v}^\top \underbrace{\left( \frac{1}{\sigma^2} + \frac{a(t)^2}{b(t)^2} \right) \mathbf{I}}_{\Lambda(t)} \mathbf{v} - \mathbf{v}^\top \underbrace{\left( \frac{\mathbf{d}_v+W^\top\mathbf{h}}{\sigma^2} + \frac{a(t)\mathbf{x}(t)}{b(t)^2} \right)}_{\boldsymbol{\eta}(\mathbf{h},\mathbf{x}(t))} + \tilde{C}.
\end{align}
This gives the conditional Gaussian density:
\begin{equation}
    q(\mathbf{v}|\mathbf{h},\mathbf{x}(t)) = \mathcal{N}\big(\mathbf{v};\, \boldsymbol{\mu}(\mathbf{h},\mathbf{x}(t)),\, \Lambda(t)^{-1}\mathbf{I}\big),
\end{equation}
with scalar precision $\Lambda(t) = \sigma^{-2} + a(t)^2 b(t)^{-2}$ and mean vector $\boldsymbol{\mu}(\mathbf{h},\mathbf{x}(t)) =  \Lambda(t)^{-1} \boldsymbol{\eta} = \Lambda(t)^{-1} \left( \frac{\mathbf{d}_v+W^\top\mathbf{h}}{\sigma^2} + \frac{a(t)\mathbf{x}(t)}{b(t)^2} \right)$, which matches Eq.~\eqref{eq:v_given_h_tethered}.

\newpage

\section{Supplementary Figures}
\begin{figure}[h]
    \centering
    \includegraphics[width=0.5\linewidth]{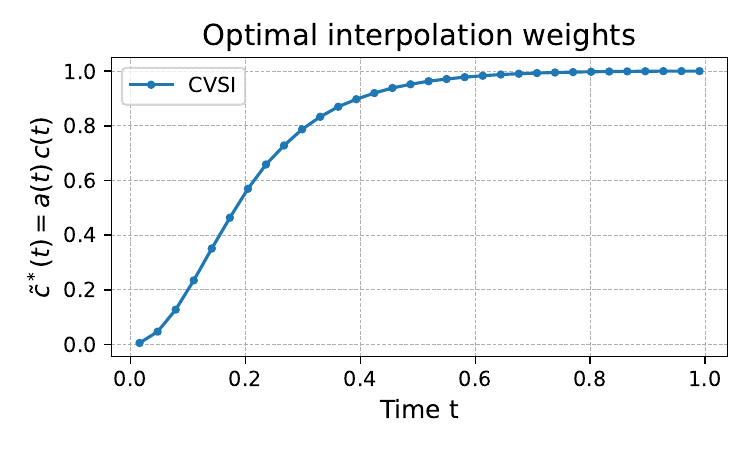}
    \caption{
    \textbf{Optimal Interpolation Weights.} The derived optimal interpolation weight $\tilde{c}^*(\mathbf{x}(t), t) = c^*(\mathbf{x}(t), t)a(t)$ (Eq.~\eqref{eq:opt_c_t_result}) smoothly transitions from 0 (TSI-dominant) to 1 (DSI-dominant), automatically selecting the stable estimator across the entire diffusion process without requiring manual tuning. The exact shape and values of $\tilde{c}^*(\mathbf{x}(t), t)$ depend on the instantaneous correlation between the target geometry and the diffusion posterior, and depends on the chosen noise schedule.}
    \label{fig:tilde_ct}
\end{figure}

\begin{figure}[h]
    \centering
    \includegraphics[width=1.\linewidth]{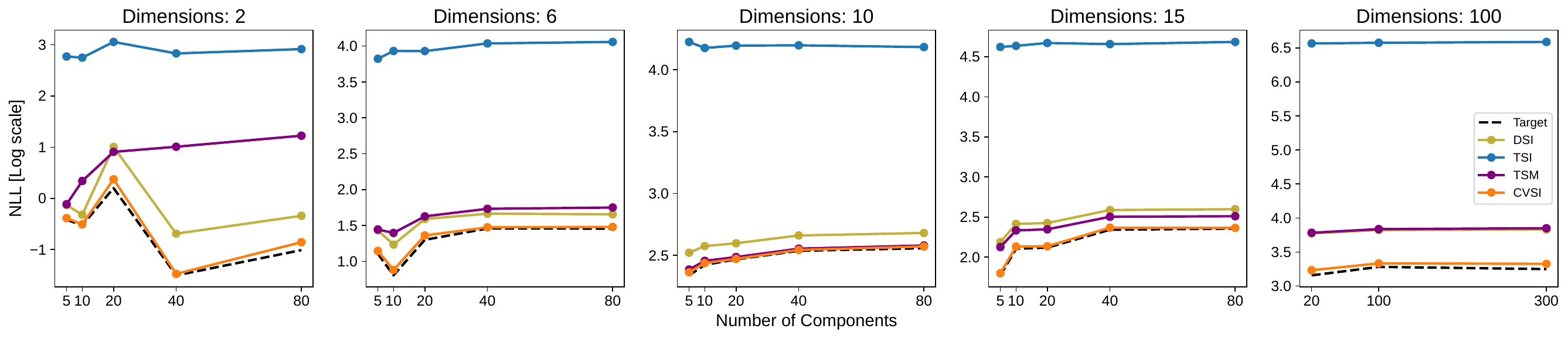}
    \includegraphics[width=.9\linewidth]{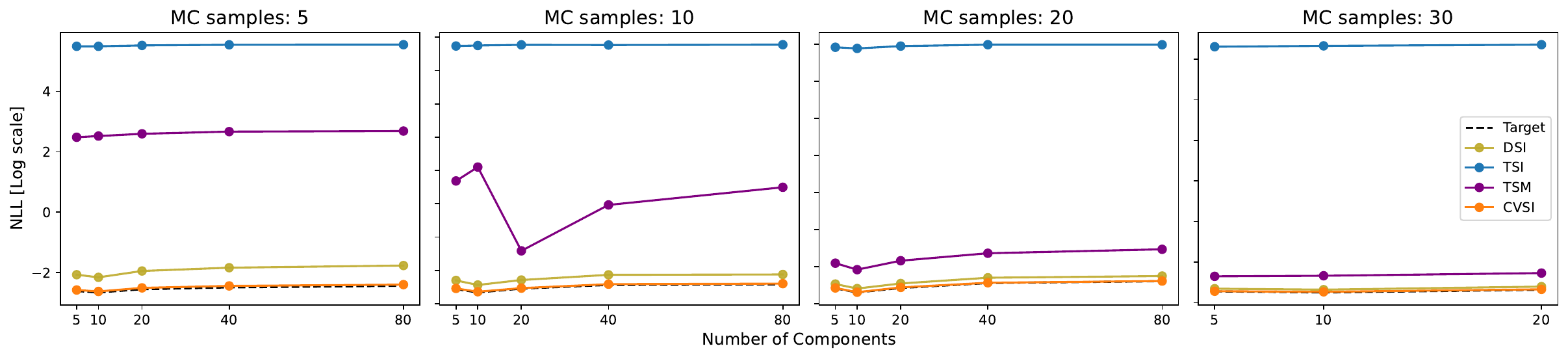}
    \caption{
    \textbf{Scalability and Sample Efficiency of Score Estimators.}
    \textit{Top:} Negative Log-Likelihood (NLL) as a function of target distribution complexity (number of GMM components) across increasing spatial dimensions ($d \in \{2, \dots, 100\}$), utilizing a fixed budget of 10 MC posterior samples per step. As dimensionality and complexity grow, baseline estimators (DSI, TSI, TSM) degrade and diverge from the ground truth. CVSI maintains optimal performance.
    \textit{Bottom:} NLL versus complexity evaluated at a fixed dimension ($d=15$) for varying MC sample budgets ($K \in \{5, 10, 20, 30\}$). CVSI achieves near-optimal NLL with as few as 5 MC samples, significantly outperforming baselines.
    }
    \label{fig:gmm_nll_vs_dim_comp}
\end{figure}

\begin{figure}[h]
    \centering
    \includegraphics[width=1.0\linewidth]{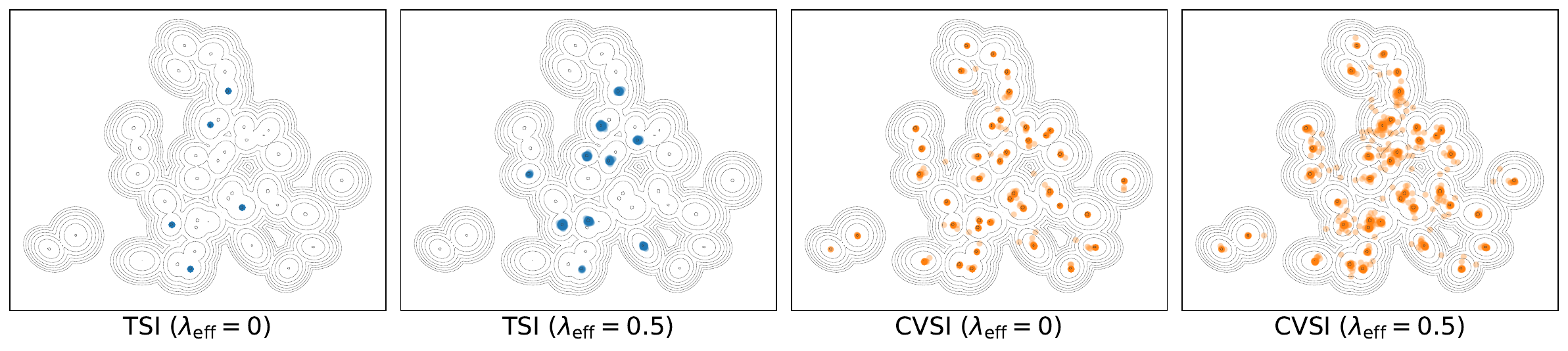}
    \caption{
    \textbf{Mitigating Temperature Inflation in High Dimensions (6D GMM).} Samples generated by iDEM for a 6-dimensional, 40-component GMM, projected onto the first two coordinates. The exact marginal ground-truth density is marked by grey contours. \textbf{Left to right:} TSI at $\lambda_\mathrm{eff}=0$ and $\lambda_\mathrm{eff}=0.5$, then CVSI at $\lambda_\mathrm{eff}=0$ and $\lambda_\mathrm{eff}=0.5$. At both levels, TSI suffers from severe mode collapse, while our CVSI recovers all the target modes.
    }
    \label{fig:gmm6d_lambda}
\end{figure}

\begin{figure}[h]
    \centering
    \includegraphics[width=1.0\linewidth]{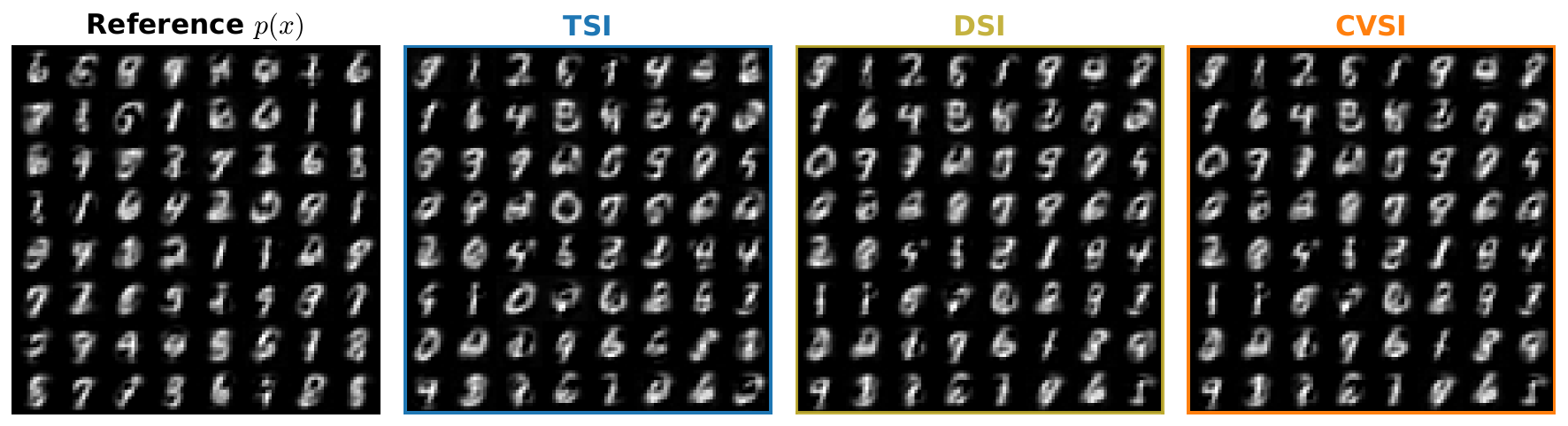}
    \caption{
    \textbf{Additional MNIST generations per estimator.} A larger sample of digits generated by simulating the reverse SDE on the Gaussian--Bernoulli RBM energy model, one panel per estimator, next to reference draws from $p(\mathbf{x})$ (panels outlined in the estimator colour). These expand the eight-sample rows of Fig.~\ref{fig:mnist_results}a.
    }
    \label{fig:mnist_samples_full}
\end{figure}

\begin{figure}[h]
    \centering
    \includegraphics[width=0.7\linewidth]{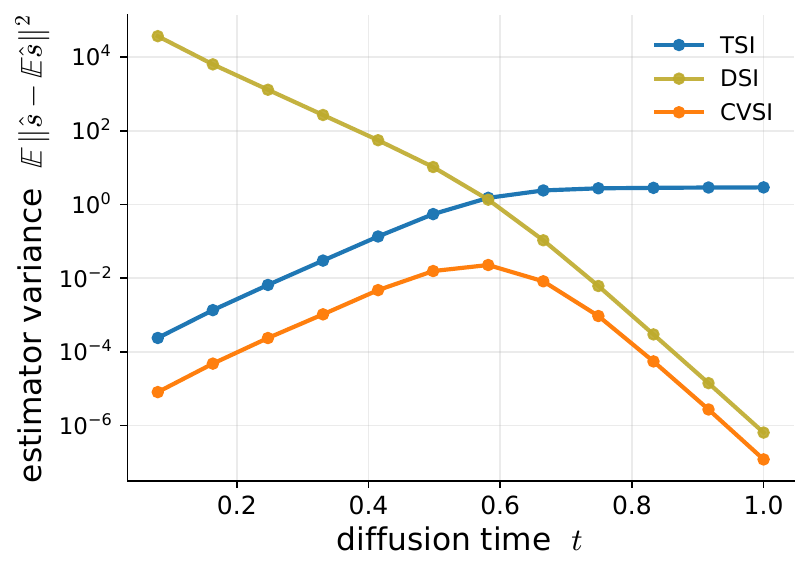}
    \caption{
    \textbf{Estimator variance on the MNIST energy model.} Monte Carlo variance of the TSI, DSI and CVSI score estimates as a function of diffusion time $t$, computed from block-Gibbs posterior draws of the GB--RBM ($D=196$). Similar to the results in Fig.~\ref{fig:control_variate_variance}, TSI diverges at high noise, DSI at low noise, and CVSI achieves the lowest variance across the entire trajectory, at no additional energy evaluation cost.
    }
    \label{fig:mnist_variance}
\end{figure}

\newpage
.

\end{appendices}

\end{document}